\documentclass[11pt,oneside]{amsart}
\title[Separation capacity of linear reservoirs]{Separation capacity of linear reservoirs with random connectivity matrix}
\date{}

\author{Youness Boutaib}
\address{Department of Mathematical Sciences\\ 
University of Liverpool\\
Liverpool, L69 7ZL. United Kingdom}
\email{Youness.Boutaib@liverpool.ac.uk}
     
\thanks{The author is grateful to E. Azmoodeh for some insightful discussions when this work was being conducted.}

\usepackage{amssymb} 
\usepackage{amsfonts}
\usepackage{amsmath} 
\usepackage{amsthm} 
\usepackage[margin=1in]{geometry}  
\usepackage[usenames,dvipsnames]{xcolor} 
\usepackage{float} 
\usepackage{graphicx} 
\usepackage{wrapfig}
\usepackage{hyperref}
\usepackage{subcaption} 
\usepackage{adjustbox} 
\usepackage{multirow}

\newtheorem{theorem}{Theorem}
\newtheorem{lemma}[theorem]{Lemma} 
\newtheorem{proposition}[theorem]{Proposition} 
\newtheorem{remark}[theorem]{Remark}

\newtheorem{definition}[theorem]{Definition}
\newtheorem{notation}[theorem]{Notation}

\newcommand{\abf}{\mathbf{a}}

\newcommand{\ibf}{\mathbf{i}}
\newcommand{\jbf}{\mathbf{j}}
\newcommand{\kbf}{\mathbf{k}}
\newcommand{\xbf}{\mathbf{x}}
\newcommand{\ybf}{\mathbf{y}}
\newcommand{\zbf}{\mathbf{z}}

\newcommand{\Ebb}{\mathbb{E}}

\newcommand{\Nbb}{\mathbb{N}}
\newcommand{\Pbb}{\mathbb{P}}

\newcommand{\Rbb}{\mathbb{R}}

\newcommand{\Ccal}{\mathcal{C}}

\newcommand{\Gcal}{\mathcal{G}}

\newcommand{\Ical}{\mathcal{I}}

\newcommand{\Lcal}{\mathcal{L}}

\newcommand{\Ncal}{\mathcal{N}}

\newcommand{\inti}[2]{[\![ #1, #2 ]\!]}

\newcommand{\opnorm}[1]{\left\| #1 \right\|_{\mathrm{op}}}

\newcommand{\rom}[1]{\uppercase\expandafter{\romannumeral #1\relax}}

\usepackage[english]{babel}

\begin{document}

\maketitle

\begin{abstract}
A natural hypothesis for the success of reservoir computing in generic tasks is the ability of the untrained reservoir to map distinct input time series to separable reservoir states, a property we term \emph{separation capacity}. In this work, we develop a rigorous mathematical framework for analysing the separation capacity of random linear reservoirs. We show that the expected separation induced by a random reservoir is completely characterised by the spectral properties of a positive semi-definite matrix naturally associated with the connectivity matrix, which we call the \emph{generalised matrix of moments}. 

Taking Gaussian connectivity matrices as an example, we investigate how separation depends on the reservoir dimension $N$, the scaling of the connectivity matrix, and structural assumptions such as symmetry. In the symmetric case, we show that, although the classical scaling $N^{-1/2}$ yields the most balanced separation for large reservoirs, the quality of separation inevitably deteriorates as the length of the input time series increases. In contrast, for reservoirs with independent and identically distributed connectivity entries, we prove that the classical scaling $N^{-1/2}$ is asymptotically optimal from the perspective of separation and derive quantitative bounds describing the evolution of separation with the time horizon.
	
	Finally, numerical experiments suggest a strong practical connection between balanced separation profiles and downstream learning performance. Beyond providing a theoretical justification for several common reservoir design choices, our results introduce separation capacity as a tractable and informative framework for the analysis of random reservoirs.
\end{abstract}

\section{Introduction}
	\subsection{The reservoir computing paradigm}\label{subsec:RCParadigm}
	Recurrent neural networks (RNNs) were one of the earliest machine learning architectures specifically tailored for processing sequential data streams. They are able to track long term temporal dependencies and to handle variable-length data, such as financial time series or text. In contrast to feed-forward neural networks (FFNN), RNNs are composed of relatively few neurons connected in a strongly recurrent manner. The building block of RNNs is mapping input time series (of any length) to a hidden state of fixed length, which is then fed to another parametric map (e.g., a regression map or a classifier.) More formally, given a $d$-dimensional time series $\mathbf{x}=(x_t)_{0\leq t \leq T} \in (\Rbb^d)^{(T+1)}$, the output $\mathbf{y}=(y_t)_{0\leq t \leq T} \in (\Rbb^n)^{(T+1)}$ of a basic RNN is computed sequentially as follows
	\begin{equation}\label{eq:RNNSysEq}
	\left\{\begin{array}{ll}
	z_0 = \phi(u x_0) \in \Rbb^N,\\
	z_t = \phi(W z_{t-1} + ux_t) \in \Rbb^N,&\text{for all } 1\leq t\leq T,\\
	y_t = h(z_t) \in \Rbb^n,\\
	\end{array}\right.
	\end{equation}	 
	where $u\in \Rbb^{N\times d}$ is a pre-processing matrix (sometimes called input mask), $W\in \Rbb^{N \times N}$, called the connectivity matrix, models the connection strength between neurons, $\phi: \Rbb \to \Rbb$ is an activation function applied element-wise and $h$ is a parametric map (e.g. a hyperplane classifier or a standard FFNN.) In principle, and very much like in a classical FFNN, all the parameters $u$, $W$ and $h$ can be optimised, using a gradient descent algorithm, by minimising a suitable empirical risk function applied to a training dataset. However, it is shown both theoretically and practically that such algorithms may either fail to converge, or do converge toward a saddle point (\cite{Doya, BFS, LHL}). In particular, and due to the recurrent nature of the architecture, the final hidden state $z_T$ involves $T-t$ applications of the connectivity matrix $W$ on the past input value $x_t$, for all $t\in \inti{0}{T}$. Thus, when handling very long time series, optimising $W$ can easily lead to numerical instability problems and a terminal state of the network that fails to appropriately take into account the contribution of the early terms of the data streams.
	
	Two of the most successful innovations to improving the performance of RNNs were the introduction of gating (e.g. Long Short-Term Memory networks (LSTMs, \cite{HS}) and Gated Recurrent Units (GRUs, \cite{JGPSTSB})) and attention (e.g. Transformers, \cite{VSPUJGKP}.) However, the well documented empirical success of such techniques comes at a high computational cost and, most importantly, a lack of theoretical analysis and guarantees and adequate interpretation. 
	
	An alternative approach, reservoir computing (RC), proposed in \cite{JH} and \cite{MJS}, offers a simpler yet effective solution. In RC, the connectivity matrix $W$ is randomly chosen rather than learned. In this context, the map $\mathbf{x}\mapsto \mathbf{z}=(z_t)_{0\leq t \leq T}$ is called a reservoir, $W$ the reservoir matrix, and $N$ the reservoir dimension. Typically, $N$ is chosen to be very large and $W$ to have either its largest singular value or its spectral radius close to but strictly smaller than 1. This type of networks is much easier to train: only $h$, and sometimes $u$, are learned; in some implementations, e.g. \cite{DCREA}, even $u$ is randomly generated. In practice, reservoir computers have shown excellent performances (\cite{JH, SH, FDHGJTLH, GTSAH, SFC}) and are even implemented in hardware (e.g.\ \cite{ASVDMDSMF, LKDB, VMVFMVSDB, CZCQLWL}) due to their multi-tasking abilities (since only few parameters need to be retrained in function of the task at hand.) In particular, the survey \cite{TYHNKTNNH} provides a review of some of the physical implementations of this paradigm.
	\subsection{Mathematical foundations of reservoir computing and current limitations}
	Despite its empirical success, a comprehensive mathematical explanation for the effectiveness of RC remains elusive. Nevertheless, several important theoretical directions have emerged in recent years and provide useful context for the present work.
	
	A first line of research concerns the approximation properties of reservoir systems. Under various assumptions on the input space and the reservoir dynamics, several universality results have shown that RC systems can approximate broad classes of dynamical systems and filters (e.g.\ \cite{BC,GO,GO2,GGO}).
	
	A second line of research seeks to understand how information from the past is encoded in the reservoir state. Using the Cayley-Hamilton theorem, the authors of \cite{Tino,VALT} represent the reservoir state as a feature map acting on a finite-dimensional \emph{network encoded input}. In special cases, such as cyclic reservoirs, this representation can be described explicitly, whereas for random connectivity matrices it is mostly understood through numerical experiments. Closely related is the notion of memory capacity (\cite{Jaeger,GHS,GGO3,CWSA}), which quantifies the ability of a reservoir to retain and reconstruct past inputs. For linear reservoirs, memory capacity can be expressed in terms of the rank of the associated controllability matrix (\cite{Jaeger,GGO3}). While these results provide valuable insight, they typically rely on assumptions on the input distribution and on the covariance structure of the reservoir states, conditions that are often difficult to verify in practice. Hence, their direct practical exploitation for reservoir design remains limited.
	
	A third perspective is provided by statistical field theory and mean-field approaches. In particular, the book \cite{MD} offers a theoretical justification for several common design choices in reservoir computing. Among these is the widespread practice of choosing the entries of the connectivity matrix $W$ to be centred random variables with variance of order $N^{-1}$, where $N$ denotes the reservoir dimension. The same reference motivates the use of large reservoirs for interpretability reasons, as this is where mean-field approximations become increasingly accurate.
	
	Despite these advances, the practical design of random reservoirs is still largely guided by heuristics, while many aspects of its empirical success remain unexplained by the existing theories.
	For example, why does the classical scaling $N^{-1/2}$ perform so well in practice? 
	What difference should one expect between symmetric random matrices and matrices with independent entries? 
	To what extent does the reservoir dimension influence the quality of the resulting representation? 
	In several previous attempts to heuristically address these questions, deterministic properties of $W$ are effectively inferred from asymptotic average behaviour (for example through expected spectral radii or operator norms), which can be a misleading intuition. As a consequence, the impact of the distribution of $W$ and of the reservoir dimension on its behaviour remains poorly understood. 
	
	The present work addresses this gap by studying random reservoirs through the lens of separation capacity, a quantity that can be analysed directly as a function of the distribution, scaling, and structure of the connectivity matrix. Rather than focusing on approximation properties, memory reconstruction, or mean-field limits, we investigate how these characteristics influence the ability of a reservoir to separate input time series.
	\subsection{Understanding reservoir computing through separation capacity} 
	In this study, we aim to introduce and explore the notion of separation capacity as  a rigorous mathematical framework for understanding the success of RC. By separation capacity, we mean the ability of an untrained reservoir to map distinct input time series to separable reservoir states. Of high interest to us, and as a guiding motivation, is the case where the connectivity matrix is randomly generated according to a prescribed structure and distribution, as this remains one of the least understood theoretical aspects of RC.
	
	The motivation behind the theory is simple. In RC, the reservoir itself is typically generated independently of the task at hand and remains untrained, while only the output map $h$ in \eqref{eq:RNNSysEq} is learned. Consequently, if two distinct input sequences are mapped to the same reservoir state ($z$ in (\ref{eq:RNNSysEq})), then no choice of output layer can subsequently recover the distinction between them, in case that is required for the task at hand. Conversely, a reservoir that produces sufficiently separated representations provides a richer set of features from which a sufficiently expressive output layer can learn (as suggested by classical universal approximation theorems). Our objective is therefore to quantify this notion of separation (by metrics that can either be precisely computed or estimated) and to understand how it depends on the distribution and structure of the random connectivity matrix. It is worth noting that our use of the term separation capacity differs from its meaning in \cite{DGJS}, where it refers to the ability of a FFNN to map separable sets into linearly separable ones. 
	
	Let us also remind that RC (with a randomly generated connectivity matrix $W$) typically produces state-of-the-art results in tasks where extracting meaningful, task-specific patterns from time series is inherently challenging. This is usually the case with short time-series or highly oscillatory, random or chaotic systems such as those studied in the references cited in Subsection \ref{subsec:RCParadigm} (financial data, pandemic hospitalisation numbers, chaotic attractors, etc.) In contrast, when more effective methods for processing time series or extracting relevant patterns are available, then other popular techniques usually perform better (for instance, transformer models for natural language processing.) In this paper, we focus on the \emph{task-independent} aspect of RC and  study the separation capacity as a necessary condition for the success of RC in \emph{generic tasks} (since the generation of the random connectivity matrix is task-independent.) Consequently, separation should not be viewed as a universal objective for learning, but rather as a task-independent property of the reservoir itself. The viewpoint adopted here is that understanding the separation profile of the untrained reservoir is an important step toward understanding the success of reservoir computing despite its reliance on task-independent parameters. From a practical point of view, when it is easy to extract data patterns or when having a very specific task in mind (such as classification), one may need to use the concept of separation in a more nuanced and task-specific way and consider the full learning process (potentially taking into consideration some properties of the data generating distribution as is common in the RC literature), as forcing full separation in this case might not be conducive to efficient learning.
	
	The relationship between separation and performance naturally raises questions about generalisation. While our work does not establish a direct bound on generalisation errors, we refer to prior work in \cite{MLB} where the authors link separation capacities of systems (such as reservoirs) to their generalisation error. More specifically, the paper introduces an algebraic notion of separation capacity, which is different but more difficult to achieve than the analytical counterpart discussed in this paper, and links it to the generalisation error, assuming the output layer to be a linear transformation of the reservoir state. While we will not investigate this question in this paper, this result offers an alternative view for the necessity of separation for performance in generic tasks. Our focus here is deliberately narrower: we study the intrinsic properties of the reservoir and how these depend on the distribution of the connectivity matrix, without imposing assumptions on the output map, the loss function, or the data-generating process. A detailed analysis of the interaction between separation capacity and generalisation therefore lies beyond the scope of the present work and constitutes an interesting direction for future research.
	
	Finally, let us note that separation capacity complements, rather than replaces, existing notions such as memory capacity. Unlike memory capacity, which is typically defined under specific assumptions on the input process and the covariance structure of the reservoir states, separation capacity can be formulated independently of any probabilistic model for the input time series. This makes it particularly well suited to the study of random reservoirs in a generic and task-independent setting.
	\subsection{Contributions and organisation of the paper}\label{subsec:Contribution}
	The primary objective of this work is to develop a mathematical theory of the separation capacity of random linear reservoirs. More specifically, we seek computable, task-independent and input-independent quantities that measure the ability of a random reservoir to map distinct input time series to well-separated reservoir states. Our main focus is the practically relevant but theoretically elusive setting in which the connectivity matrix is sampled from a prescribed probability distribution.
	
	To make the analysis tractable while preserving the essential mechanisms, we consider linear reservoirs, that is, we take the activation function $\phi$ in \eqref{eq:RNNSysEq} to be the identity. This simplification still retains much of the expressive power of the architecture. Indeed, linear reservoirs equipped with sufficiently rich readout maps remain universal approximators in suitable settings (\cite[Corollary~11]{GO3}). We also restrict our attention to one-dimensional input sequences and to the pre-processing vector $u=(1,\ldots,1)^\top$. As explained later in Remark~\ref{rem:GenUandD}, these assumptions mainly simplify the notation and the presentation; the underlying analysis technique extends to more general settings. 
	
	Our main contribution is to show that, under these assumptions, the expected separation properties of a random reservoir are completely characterised by a positive semi-definite matrix naturally associated with the connectivity matrix, which we call the \emph{generalised matrix of moments}. Consequently, the study of separation capacity reduces to a spectral problem. This connection allows us to bring tools from random matrix theory into the analysis of random reservoirs.
	
	Focusing on the practically important Gaussian setting (although many of the techniques extend beyond the Gaussian setting, and several asymptotic arguments carry over to broader classes of sub-Gaussian distributions), we investigate how separation depends on the reservoir dimension, the scaling of the connectivity matrix, and structural assumptions such as symmetry. In particular, we analyse and compare two of the most common random reservoir models: reservoirs with i.i.d.\ Gaussian connectivity matrices and reservoirs with symmetric Gaussian connectivity matrices. Our results provide explicit asymptotic descriptions in the regimes of large reservoir dimension and long time series, and offer a theoretical explanation for several heuristics commonly used in RC, including the choice of large i.i.d.\ connectivity matrices and the classical scaling of their entries by $N^{-1/2}$.
	
	More explicitly, we will proceed in the paper in the following manner. \textbf{Section \ref{sec:Notations}} introduces the mathematical model, general assumptions, notations and some preliminary remarks. 
	In \textbf{Section \ref{sec:ExpectDist}}, we examine the expected squared distance between reservoir states associated with two time series and prove that this expected separation capacity is fully encoded by the spectrum of the generalised matrix of moments (Theorem~\ref{theo:FundNdimSep}). Consequently, we introduce spectral measures of separation capacity and quality to be analysed in the following subsections. In the symmetric case (Subsection~\ref{subsec:ExpSym}), we establish asymptotic descriptions of the separation capacity and quality and show that, although the classical scaling $N^{-1/2}$ yields the most balanced separation for large reservoirs, the quality of separation eventually deteriorates as the time horizon grows (Theorems~\ref{Thm:NinftyGaussSym} and~\ref{Thm:TinftyGaussSym}).
	In the i.i.d.\ case (Subsection~\ref{subsec:ExpIID}), we show that the critical scaling $N^{-1/2}$ emerges as the asymptotically optimal scaling from the perspective of separation quality (Theorem~\ref{Thm:NinftyGaussIID}). We also obtain quantitative bounds for very long time series suggesting that scalings very far from $N^{-1/2}$ result in a rapid deterioration of separability with time (Theorem~\ref{Thm:TinftyGaussIID}). \textbf{Section~\ref{sec:numerics}} provides numerical illustrations on classification, memory, and forecasting tasks. While these experiments are not intended as performance benchmarks, they consistently suggest that reservoirs exhibiting a more balanced separation profile produce more useful representations for downstream learning tasks. Finally, \textbf{Section~\ref{sec:OProblems}} discusses several open questions, including the role of eigenvectors of the generalised matrix of moments, high-probability separation guarantees, nonlinear reservoirs, and possible links between separation capacity, memory, and generalisation.
	\section{Notations and general remarks}\label{sec:Notations}
	$\Nbb$ (respectively $\Nbb^*$) denotes the set of positive integers (resp.\ strictly positive integers.) Given $d\in \Nbb^*$, consider $d$-dimensional input time series 
	\[
	\mathbf{x}=(x_t)_{0\leq t \leq T} \in (\Rbb^d)^{(T+1)} = (x_0, x_1, \ldots, x_T),
	\] 
	where, for $0\leq t \leq T$, $x_t$ is the input received at time $t$. At each time $t$, the signal $x_t$ is transformed into an $N$-dimensional vector by a pre-processing matrix $u \in \Rbb^{N \times d}$, with $N\in \Nbb^*$. The linear reservoir with dimension $N$ and (a possibly random) connectivity matrix $W\in \Rbb^{N \times N}$ associates a reservoir state (or hidden state) $f_t(\mathbf{x}, W)\in \Rbb^N$ to such a time series computed recursively as follows
	\begin{equation}\label{eq:RCLinEq}
	\left\{\begin{array}{ll}
	f_0(\mathbf{x}, W) = ux_0 \in \Rbb^N,\\
	f_t(\mathbf{x}, W) = W f_{t-1}(\mathbf{x}, W) + ux_t \in \Rbb^N,&\text{for all } 1\leq t\leq T.\\
	\end{array}\right.	
	\end{equation}
	Note that, at time $t$, the reservoir state $f_t(\mathbf{x}, W)$ is given by the simple formula
	\[
	f_t(\mathbf{x}, W) 
	= ux_{t} +  W ux_{t-1} + \cdots +  W^t ux_{0}
	= \sum_{s=0}^t W^s ux_{t-s}.
	\]
	In particular, $f_t(\cdot, W)$ is linear. Hence, comparing the distance of the reservoir states $\|f_t(\mathbf{x}, W)  -  f_t(\mathbf{y}, W)\|_2$ to the Euclidean distance $\|\mathbf{x}-\mathbf{y}\|_2$, where $\mathbf{x}$ and $\mathbf{y}$ are two given input time series, is equivalent to understanding the link between  $\|f_t(\mathbf{a}, W)\|_2$ and $\|\mathbf{a}\|_2$ for any given input time series $\mathbf{a}$: one needs only to consider $\mathbf{a}:=\mathbf{x}-\mathbf{y}$.
	
	Next, we observe that the intermediate reservoir state $f_t(\mathbf{x}, W)$ associated with a time series $\mathbf{x}$ at time $t$ can be written as the terminal reservoir state at time $T$ associated with the delayed time series $\tau_t(\mathbf{x})$ where the first $T-t$ input values are equal to $0$, i.e.\
	\[
	 \tau_t(\mathbf{x}):=(0,\ldots, 0,x_{0}, x_{1},\ldots, x_{t}).
	\]
	More precisely
	\[\begin{array}{rcl}
	f_t(\mathbf{x}, W) 
		&=& ux_{t} +  W ux_{t-1} + \cdots +  W^t ux_{0}\\
	&=& ux_{t} +  W ux_{t-1} + \cdots +  W^t ux_{0} + W^{t+1} u\cdot 0 + \cdots + W^Tu\cdot 0\\
	&=&f_T((0,\ldots, 0,x_{0}, x_{1},\ldots, x_{t}), W) .\\
	\end{array}
	\]
	Consequently, understanding the variation in time in reservoir states $f_{t+1}(\mathbf{x}, W) - f_t(\mathbf{x}, W)$ for a single time series $\mathbf{x}$ is equivalent to understanding the difference in terminal reservoir states
	\[
	f_{T}(\tau_{t+1}(\mathbf{x}), W) - f_T(\tau_t(\mathbf{x}), W)
	\]
	associated to the delayed time series $\tau_{t+1}(\mathbf{x})$ and $\tau_t(\mathbf{x})$. In light of these arguments, we shall restrict ourselves in this paper to studying the separation properties of the terminal reservoir state map $f:=f_T$.
	
	Since we are interested in the separation capacity of a reservoir when looking at time series as a vector with entries given by their values through time, and as we are not considering the $d$-dimensional spatial structure of the signals then, without loss of generality, we will limit ourselves to one-dimensional input time series (i.e.\ $d=1$) and will assume that $u=(1,\ldots,1)^{\top}$. This is for instance in line with the setting in \cite{VALT, Tino}. This assumption considerably simplifies the exposition while preserving all the key phenomena studied in this work. The techniques developed throughout the paper extend with only minor modifications to arbitrary input dimension $d$ and pre-processing matrix $u$. We shall return to this point more explicitly in Remark~\ref{rem:GenUandD}.	

	Let us end this section with further notation that is key in this paper. For two integers $p\leq q$, $\inti{p}{q}$ denotes the set $\{p, p+1, \ldots, q\}$. Given $N\in \Nbb^*$, we denote $[N]=\inti{1}{N}$. Given a vector $x=(x_1,\ldots, x_d)\in \Rbb^d$, $\|x\|_2$ denotes its Euclidean norm, i.e.\
		\[
		\|x\|_2^2 = \sum_{i=1}^d x_i^2.
		\]
	$A^\top$ denotes the transpose of a given matrix $A$. Given $m,n \in \Nbb^*$ and a matrix $A\in \Rbb^{m\times n}$, $\opnorm{A}$ denotes the operator norm of $A$
	\[
	\opnorm{A} = \max_{x\in \Rbb^n\setminus\{0\}} \frac{\|Ax\|_2}{\|x\|_2},
	\]
	while $\|A\|_{2,\infty}$ denotes the following induced norm
	\[
	\|A\|_{2,\infty} = \max_{1\leq i\leq m} \|A_{i \cdot}\|_2
	= \max_{1\leq i\leq m}\left( 
	\sum_{1\leq j \leq n} A_{ij}^2
	\right)^{1/2}.
	\]
	Given $n \in \Nbb^*$ and a matrix $A\in \Rbb^{n\times n}$, $\mathrm{tr}(A)$ denotes the trace of $A$. When no confusion is possible, $|X|$ denotes the number of elements in a given finite set $X$. Given two strictly positive sequences $(x_t)_{t\in \Nbb}$ and $(y_t)_{t\in \Nbb}$, we write $x_T \underset{T\to +\infty}{\simeq} y_T$ if $\left(\frac{x_t}{y_t} \right)_{t\in \Nbb}$ converges to 1 at infinity. Finally, $X\sim \Ncal(m,\sigma^2)$ means that $X$ is a Gaussian random variable with mean $m$ and variance $\sigma^2$.
\section{Expected distance distortion}\label{sec:ExpectDist}
	\subsection{Generalities}
	In this section we investigate the expected distance distortion induced by a random linear reservoir. More precisely, given a one-dimensional input time series $\mathbf{a}\in\Rbb^{T+1}$ and a random connectivity matrix $W$, we study the quantity 
	\[ \frac{\|f(\mathbf{a},W)\|_2^2}{\|\mathbf{a}\|_2^2}, \] 
	which measures the amplification or contraction of Euclidean distances by the reservoir dynamics. Our goal is to understand how this separation capacity depends on the main parameters of the model: the reservoir dimension $N$, the time horizon $T$, the generating distribution of the connectivity matrix, and possible structural assumptions such as symmetry. Particular attention will be devoted to the asymptotic regimes $T\to\infty$ and $N\to\infty$.
	
	We recall that the final hidden state of the $N$-dimensional reservoir with pre-processing matrix $u$ and connectivity matrix $W$ is given by
	\[
	f(\abf,W)= \sum_{t=0}^T W^t ua_{T-t}
	=(f_i)_{1\leq i \leq N} \in \Rbb^N.
	\]
	
	As discussed in Subsection \ref{subsec:Contribution} and Section \ref{sec:Notations}, we will take $u=(1,1,\ldots,1)^\top\in \Rbb^N$. These restrictions on the dimension $d$ of the input signal and the form of $u$ are only superficial as will be discussed in details in Remark \ref{rem:GenUandD}. In particular, if need be, all our computations in the remaining of this article can be generalised, albeit with heavier notations and computations.
	
	The next theorem shows that the expected separation properties of the linear reservoir are completely determined by the spectral properties of a positive semi-definite generalised matrix of moments. This observation forms the starting point of all subsequent developments.
	
	\begin{theorem}\label{theo:FundNdimSep} Consider an $N$-dimensional linear reservoir with random connectivity matrix $W$, i.e. the output of the reservoir for the signal $\xbf:=(x_t)_{0\leq t \leq T}$ is given by 
	\[
	f(\xbf,w)= \sum_{t=0}^T W^t ux_{T-t},
	\]
	where $u=(1,1,\ldots,1)^\top$. Then the expected separation capacity of the reservoir is characterised by the eigenvalues of the symmetric positive semi-definite matrix
	\begin{equation}\label{eq:GenHankelMatrix}
	B_{T,N}:= \Ebb \left(\sum_{i=1}^N\beta_{i,l_1}\beta_{i,l_2}\right)_{0 \leq l_1,l_2 \leq T} \in \Rbb^{(T+1)\times (T+1)},
	\end{equation}
	where, for $i\in \inti{1}{N}$
	\[
	\beta_{i,0}= 1 \quad \text{and}\quad
	\beta_{i,l}= 
	\sum_{i_1,i_2, \ldots, i_{l}=1}^N W_{ii_1}W_{i_1i_2}\cdots W_{i_{l-1} i_l}.
	\]
	More specifically, if $\lambda_{\mathrm{min}} (B_{T,N})$ and $\lambda_{\mathrm{max}} (B_{T,N})$ denote respectively the smallest and largest eigenvalue of $B_{T,N}$ and given two time series $\xbf$ and $\ybf$, one has 
	\begin{equation}\label{eq:ExpSepNdim}
	\lambda_{\mathrm{min}} (B_{T,N}) \|\xbf-\ybf\|^2_2 \leq 
	\Ebb \|f(\xbf,W)-f(\ybf,W)\|^2_2 \leq \lambda_{\mathrm{max}} (B_{T,N}) \|\xbf-\ybf\|^2_2.
	\end{equation}
	\end{theorem}
	
	\begin{proof}
	Given two time series $\xbf$ and $\ybf$, and defining $\abf:=\xbf-\ybf$, we get $f(\xbf,W)-f(\ybf,W)=f(\abf,W)$. Denoting $f(\abf,W)=(f_i)_{1\leq i \leq N} \in \Rbb^N$, then, for all $i\in \inti{1}{N}$
	\[
	f_i=a_T
	+ \left(\sum\limits_{j=1}^N W_{ij}\right) a_{T-1}
	+\cdots
	+ \left(\sum\limits_{i_1,i_2, \ldots, i_{T}=1}^N W_{ii_1}W_{i_1i_2}\cdots W_{i_{T-1} i_T}\right) a_{0}= \sum\limits_{l=0}^T a_{T-l}\beta_{i,l}.
	\]
	 Hence	
	\[
	\|f(\abf,W)\|^2_2 = \sum_{i=1}^N f_i^2 
	=  \sum_{i=1}^N \left(\sum_{l=0}^T a_{T-l}\beta_{i,l} \right)^2
	=  \sum_{l_1,l_2=0}^T a_{T-l_1}a_{T-l_2}\left(\sum_{i=1}^N\beta_{i,l_1}\beta_{i,l_2} \right).
	\]
	Therefore $\Ebb \|f(\abf,W)\|^2_2 = a^\top B_{T,N} a$, where $a=(a_T,\ldots,a_0)^\top$, which then yields Inequality (\ref{eq:ExpSepNdim}).
	\end{proof}
	
	\begin{remark}\label{rem:GenUandD} In the general case where the dimension $d$ of the input signal and the pre-processing map $u\in \Rbb^{N \times d}$ are arbitrary, one can write
	\[
	f(\abf,W) = W_{T,N} U A,
	\]
	where
	\[
	W_{T,N} =
	\left[
	\mathrm{I}_N, W, \cdots, W^T
	\right] \in \Rbb^{N\times N(T+1)},
	\]
	\[
	U=\mathrm{diag}(u, \ldots, u)
	=\left[ \begin{array}{cccc}
	u & 0_{N\times d} & \cdots & 0_{N\times d}\\
	0_{N\times d} & u & \cdots & 0_{N\times d}\\
	\vdots & \vdots  &\ddots &\vdots \\
	0_{N\times d} & 0_{N\times d} & \cdots & u\\
	\end{array}
	\right]
	\in \Rbb^{N(T+1) \times d(T+1)},
	\]
	and
	\[
	A = \left[ \begin{array}{c}
	a_T\\
	a_{T-1}\\
	\vdots \\
	a_0\\
	\end{array}\right]\in \Rbb^{d(T+1)}.
	\]
	This yields a formula that is similar to the one obtained in the proof of Theorem \ref{theo:FundNdimSep}
	\[
	\Ebb \|f(\abf,W)\|^2_2 = A^\top U^\top W_{T,N}^\top W_{T,N} U A.
	\]
	
	One can choose to lead the subsequent analysis in one of two possible manners. The first approach is to understand the direct impact of the distribution of $W$ on separation, independently of the choice of the pre-processing map $u$, i.e.\ directly treating $(u(a_t))_{0\leq t \leq T}$ as the input signal. In this case, we write
	\[
	\Ebb \|f(\abf,W)\|^2_2 = \tilde{a}^\top B^*_{T,N} \tilde{a},
	\]
	where $\tilde{a} \in \Rbb^{N(T+1)}$ is the vectorised processed input
	\[
	\tilde{a}=UA=
	\left(
	\begin{array}{c}
	u a_T\\
	\vdots \\
	u a_0\\
	\end{array}
	\right)=
	\left(
	\begin{array}{c}
	(u a_T)_1\\
	\vdots \\
	(u a_T)_N\\
	\vdots \\
	(u a_0)_1\\
	\vdots \\
	(u a_0)_N\\
	\end{array}
	\right)
	,
	\]
	and the positive semi-definite matrix $B^*_{T,N}= \Ebb W_{T,N}^\top W_{T,N} \in \Rbb^{N(T+1)\times N(T+1)}$ is given by
	\[
	\forall 0\leq l_1,l_2 \leq T,\;\;
	\forall 1 \leq j_1, j_2 \leq N:\quad
	(B^*_{T,N})_{l_1 N + j_1, l_2 N + j_2} = \Ebb \sum_{i=1}^N (W^{l_1})_{i,j_1}(W^{l_2})_{i,j_2}.
	\]
	The second approach, which is similar to what we will carry out in the remaining of this article with $d=1$ and $u=(1,\ldots,1)^\top$, is to directly include the effect of $u$ into the analysis of the separation capacity. In this case, we write
	\[ \Ebb \|f(\abf,W)\|^2_2 =  A^\top B_{U,T,N} A,\]
	where the positive semi-definite matrix $B_{U,T,N}=U^\top B^*_{T,N} U \in \Rbb^{d(T+1)\times d(T+1)}$ is explicitly given by, for all $0\leq l_1,l_2 \leq T$ and $1 \leq j_1, j_2 \leq d$ 
		\[
	(B_{U,T,N})_{l_1 d + j_1, l_2 d + j_2} 
	= \sum_{1\leq k_1, k_2 \leq N} u_{k_1, j_1}u_{k_2, j_2} \Ebb \sum_{i=1}^N (W^{l_1})_{i,k_1}(W^{l_2})_{i,k_2}.
	\]	
	In other words, the entries of $B_{U,T,N}$ are local reweightings of the entries of $B^*_{T,N}$. In a matrix form, the effect of the pre-processing map $u$ is given by the effect of the matrix $U$ on the eigenvalues of $B^*_{T,N}$ through the transform $U^\top B^*_{T,N}U$.
	
	In both cases, it should be very clear in the sequel that the computations can be carried out very similarly.
	\end{remark}
	
	\begin{remark}\label{rem:LinkMemoryControl} Using the notation of Remark \ref{rem:GenUandD}, let us make the following links with other classical objects in reservoir computing analysis.
	
	In the special case $T=N-1$, the matrix 
	\begin{equation}\label{eq:ControlMatrix}
	\Ccal=W_{N-1,N} U 
	= 	\left[
	u, Wu, \cdots, W^{N-1}u
	\right],	
	\end{equation}
	is known as the controllability matrix of the reservoir. A fundamental result in RC states that the memory capacity of a linear reservoir with independent inputs is given by the rank of $\Ccal$ (\cite{Jaeger, GGO3}.) \cite{RT} uses this to make a case for not training $u$: choosing $u$ with aperiodic entry sign maximises the memory capacity of reservoirs with a cyclic connectivity matrix.
	
	The matrix $B_{U,T,N}$ is sometimes called the metric tensor of the reservoir kernel. It has been used in several studies of linear reservoirs viewed as kernel machines; for example, to quantify the relative contribution of recent and distant inputs (\cite{Tino,VALT}), or to show that some reservoirs with cyclic connectivity matrix separate reservoir states according to their Fourier transforms (\cite{FLT}).
	\end{remark}
	
	\begin{remark}\label{rem:HankelDim1} For $N=1$, $B_{T,1}= (\Ebb W^{i+j})_{0\leq i,j \leq T}$ is sometimes called the matrix of moments. It has the structure of what is called a Hankel matrix (i.e., $B_{T,1}(i,j)=B_{T,1}(k,l)$, whenever $i+j=k+l$.) Moment matrices play a central role in the classical moment problem, where one seeks to determine whether a probability distribution is uniquely characterised by its moments (see, e.g., \cite{Hamburger,ST,BCI}). In contrast, for $N\geq 2$, $B_{T,N}$ no longer possesses a Hankel structure in general. For instance, if $W\in \Rbb^{2\times 2}$ has i.i.d.\ standard Gaussian entries, then it is easy to check that $B_{2,2}(0,2)\neq B_{2,2}(1,1)$. We shall see later, however, that if $W$ is symmetric, then $B_{T,N}$ is a Hankel matrix (see Remark \ref{rem:WsymBHankel}.)
	\end{remark}
	
	The full spectral decomposition of $B_{T,N}$ provides an orthonormal basis of $\Rbb^{T+1}$ that enables us to decompose time series in the eigenvalue directions and then understand which parts of the time series are best separated (in expectation). This is in contrast to the naive but popular ``recent versus distant past'' comparison. In the present paper, we will not focus on the eigenvectors of $B_{T,N}$ and their evolution in time any further.
	
	In view of its importance in the rest of this paper, let us fix the definition of $B_{T,N}$. 
	\begin{definition}\label{defn:GenHankelMatrix} Let $W\in \Rbb^{N\times N}$ be a random matrix and $T\in \Nbb^*$. We call the generalised matrix of moments of $W$ of order $T$ the matrix denoted $B_{T,N}$ that is defined by
	\[
	B_{T,N}= (B_{T,N}(l_1,l_2))_{0 \leq l_1,l_2 \leq T}:= \Ebb \left(\sum_{i=1}^N\beta_{i,l_1}\beta_{i,l_2}\right)_{0 \leq l_1,l_2 \leq T} \in \Rbb^{(T+1)\times (T+1)},
	\]
	where, for $i\in \inti{1}{N}$, $\beta_{i,0}= 1$ and
	\[
	\beta_{i,l}= 
	\sum_{i_1,i_2, \ldots, i_{l}=1}^N W_{ii_1}W_{i_1i_2}\cdots W_{i_{l-1} i_l}.
	\]
	\end{definition}
	
	Throughout this paper, we will use the following unified notation.
	\begin{notation} Given a random matrix $W\in \Rbb^{N\times N}$ and $T\in \Nbb^*$, we denote by $B_{T,N}$ its generalised matrix of moments of order $T$. 
	
	Furthermore, we denote by $\lambda_{\mathrm{max}} (B_{T,N})$ (respectively $\lambda_{\mathrm{min}} (B_{T,N})$) its largest (resp.\ smallest) eigenvalue. More generally, the eigenvalues of $B_{T,N}$ are denoted and ordered as follows
	\begin{equation}\label{eq:EVBNotation}
	(\lambda_{\mathrm{min}} (B_{T,N})=)\lambda_{T} (B_{T,N}) \leq \lambda_{T-1} (B_{T,N}) 
	\leq \cdots \leq 
	 \lambda_{0} (B_{T,N}) (=\lambda_{\mathrm{max}} (B_{T,N})).
	 \end{equation}
	\end{notation}
	
	Fix $N\in \Nbb^*$. On the one hand, Theorem \ref{theo:FundNdimSep} highlights the importance of understanding the behaviour, in function of $T$ and $N$, of the smallest eigenvalue $\lambda_{\mathrm{min}} (B_{T,N})$ as a worst-case theoretical guarantee of (expected) separation of the reservoir states associated to very long input time series or very large reservoirs. On the other hand, and from a practical viewpoint, Inequality \eqref{eq:ExpSepNdim} highlights another equally important problem: understanding the \emph{dominance} of the largest eigenvalue $\lambda_{\mathrm{max}} (B_{T,N})$ (of $B_{T,N}$) over $\lambda_{\mathrm{min}} (B_{T,N})$ or more generally over the rest of the spectrum of $B_{T,N}$. Indeed, this allows us to determine whether the separation is distributed across many directions of the input space or concentrated along only a few dominant modes. 
	
	 Combined with the difficulty of obtaining sharp estimates for $\lambda_{\mathrm{min}} (B_{T,N})$ in general, it is this latter point of view of the spectral concentration of $B_{T,N}$ that will dictate the line of reasoning in the rest of the paper. More specifically, and as a measure of the quality of separation, we introduce the dominance ratio
	\begin{equation}\label{eq:DomRatioN}
	r_{T,N} := \frac{\lambda_{\mathrm{max}} (B_{T,N})}{\sum\limits_{i=0}^T \lambda_{i} (B_{T,N})}.
	\end{equation}
	 Values of $r_{T,N}$ close to $1$ indicate that most of the expected separation is concentrated in a single eigendirection. In that regime, the reservoir effectively behaves as a nearly one-dimensional separator, irrespective of the ambient dimension of the input space.

	In the following proposition, we give some generic theoretical guarantees on the values of the largest and smallest eigenvalues of $B_{T,N}$. While these might seem basic, they have the advantages of being dimension-free and easy to compute. Moreover, we will see later that these will be enough to give us some idea about the evolution of the dominance ratio $r_{T,N}$ of $B_{T,N}$, and even capture the order of magnitude of the largest eigenvalue in many classical examples.
	\begin{proposition}\label{prop:EVboundsNdim} Let $N\in \Nbb^*$ and $W\in \Rbb^{N\times N}$ be a random matrix. Then $(\lambda_{\mathrm{max}} (B_{T,N}))_{T\geq 1}$ (respectively $(\lambda_{\mathrm{min}} (B_{T,N}))_{T\geq 1}$) is an increasing (resp.\ decreasing) sequence. Moreover,
	\begin{equation}\label{eq:LambdaIneqNDim}
	\max\limits_{0\leq l_1\leq T} 
	\sqrt{\sum\limits_{l_2=0}^T  \left( B_{T,N}(l_1,l_2)\right)^2}
	\leq \lambda_{\mathrm{max}} (B_{T,N})
	\leq \mathrm{tr}(B_{T,N})
	= \sum_{l=0}^T B_{T,N}(l,l).
	\end{equation}
	\end{proposition}
	\begin{proof}
	The monotonicity of the sequences $(\lambda_{\mathrm{max}} (B_{T,N}))_{T\geq 1}$ and $(\lambda_{\mathrm{min}} (B_{T,N}))_{T\geq 1}$ can be obtained for example using Cauchy's interlacing theorem (e.g.\ \cite{Hwang}). More specifically, one has
	\[
	\lambda_{T+1} (B_{T+1,N}) \leq \lambda_{T} (B_{T,N}) \leq \lambda_{T} (B_{T+1,N}) 
	\leq \cdots 
	\leq \lambda_{0} (B_{T,N}) \leq \lambda_{0} (B_{T+1,N}) .
	\]
	Since the eigenvalues of $B_{T,N}$ are positive and sum up to $\mathrm{tr}(B_{T,N})$, then
	\[
	\lambda_{\mathrm{max}} (B_T) \leq \mathrm{tr}(B_{T,N}).
	\]
	Finally, as the operator norm of any matrix is larger than the Euclidean norms of its rows, then
	\[
	\lambda_{\mathrm{max}} (B_{T,N})=\opnorm{B_{T,N}} \geq 
	\|B_{T,N}\|_{2,\infty}=\max_{0\leq i\leq T} \sqrt{\sum\limits_{l_2=0}^T  \left( B_{T,N}(l_1,l_2)\right)^2}.
	\]
	\end{proof}
	\begin{remark}
	In the proposition above, we have omitted giving bounds on the smallest eigenvalue of $B_{T,N}$ as basic standard algebraic results do not yield a sharp estimate of $\lambda_{\mathrm{min}} (B_{T,N})$ in general. In the case $N=1$, one can obtain such estimates following the techniques initially developed in \cite{Szego} (cf. also \cite{BCI, BS}.) For instance, if $W$ is a centred Gaussian random variable with variance $\rho^2$, then one can show that
	\[
	\lambda_{\mathrm{min}} (B_{T,1}) 
	\underset{T\to +\infty}{\simeq} 
	\frac{2^{5/2}\pi T^{1/4}}{\rho^{3/2}}
	\exp\left(\frac{1}{2\rho^2}-\frac{2 \sqrt{T}}{\rho}\right).
	\]
	However, such proof techniques are not trivial to generalise to higher dimensions.
	\end{remark}
	\begin{remark}	
	The bounds \eqref{eq:LambdaIneqNDim}, which are valid for any symmetric positive semi-definite real matrix, hold true for the general case where the dimension $d$ of the input signal and the pre-processing map $u\in \Rbb^{N \times d}$ are arbitrary (see Remark \ref{rem:GenUandD}.)
	\end{remark}
	
	In the remainder of this paper, and following the above remarks, we will try to capture the separation capacity of random linear reservoirs by understanding, on the one hand, the best-case-scenario separation capacity of the reservoir given by $\lambda_{\mathrm{max}} (B_{T,N})$, and on the other hand, the quality of this separation as captured by the dominance ratio $r_{T,N}$ (defined in \eqref{eq:DomRatioN}).
	
	Assuming sharpness of Inequality \eqref{eq:LambdaIneqNDim}, the bounds in Proposition \ref{prop:EVboundsNdim} highlight the importance of understanding the order of the terms $B_{T,N}(l_1,l_2)$ for $l_1,l_2\in \inti{0}{T}$, for $T$ or $N$ large. The methodology for the computations of such orders (especially for large $N$ and symmetric matrices) are very classical in the literature of random matrices (e.g.\ \cite{AGZ}). It is at this exact point where the simplifying assumptions $d=1$ and $u=(1,\ldots, 1)^\top$ play a role in fixing the ideas and exemplifying the computations that can be mimicked for other general cases. To simplify these computations but also to be in line with the common practical implementation of reservoirs, we will restrict ourselves in the remaining of this paper to connectivity matrices with Gaussian entries. Nevertheless, it must be clear that most computations generalise easily to other cases (e.g.\ matrices with sub-Gaussian entries.) We will in particular distinguish two cases that are common in practice: one where the connectivity matrix $W$ is assumed to be symmetric and one where all of its entries are assumed to be independent. The (random) entries of the connectivity matrix are usually chosen to have a standard deviation equal to $\frac{\rho}{\sqrt{N}}$, with $\rho$ close to $1$, and $N$ very large. For the purpose of understanding the advantage of such a choice (beyond numerical stability and easiness of interpretation), we will consider a more general class of standard deviations of the type $\frac{\rho}{N^\alpha}$, with $\alpha\geq 0$. Before we carry on, we will need a compact notation for the Gaussian moments.
	\begin{notation} Given a standard Gaussian random variable $X$, we denote its moments by
	\[
	\forall t\in \Nbb:\quad m_t:=\Ebb X^t.
	\]
	\end{notation}
	
	Trivially, for all $t,s \in \Nbb$, one has $m_t m_s \leq m_{t+s}$. Let us also introduce the following multi-index notation.
	\begin{notation}
	Given a matrix $W\in \Rbb^{N\times N}$, $l\geq 2$ and $\ibf=(i_1, i_2, \ldots, i_l) \in [N]^{l}$,
	we denote
	\[
	W_{\ibf} := W_{i_1i_2}W_{i_2i_3}\cdots W_{i_{l-1} i_{l}}.
	\]
	\end{notation}
	\subsection{The symmetric case}\label{subsec:ExpSym}
	We start with the symmetric case which is the most readily available in the literature of random matrices. Indeed, estimating the moments appearing in Definition~\ref{defn:GenHankelMatrix} and Inequality \eqref{eq:LambdaIneqNDim} is closely related to the combinatorial arguments underlying the celebrated Wigner semi-circle law (see, for example, \cite{Wigner,AGZ}). The key counting arguments may be naturally expressed in graph-theoretic terms. We therefore start with the following definition.
	\begin{definition}
	Given an undirected graph $G$ (with possibly self-loops), a sequence 
	\[w=(v_1,v_2,\ldots, v_l)\] 
	is called a walk on the graph $G$ of length $l-1$ if the following two conditions are satisfied:
	\begin{itemize}
		\item For all $i\in [l]$, $v_i$ is a vertex in $G$,
		\item For all $i\in [l-1]$, $\{v_i, v_{i+1}\}$ is an edge in $G$.
	\end{itemize}
	If $v_l=v_1$, we say that $w$ is a loop (on $G$).
	\end{definition}
	
	We denote the sets of vertices and edges of any graph $G$ by $V_G$ and $E_G$ respectively. Recall that by Euler's formula (\cite{Bollobas}), $|V_G|-|E_G|\leq 1$, with equality if and only if $G$ is a tree. 
	
	Let us now move to our main estimate.
	\begin{lemma}\label{lemma:OrderMagnitudeSymmetric}
	Let $N, T\in \Nbb^*$, $\rho>0$ and $\alpha \in \Rbb$. Let $W\in \Rbb^{N\times N}$ be a symmetric random matrix such that its entries on and above the diagonal are i.i.d.\ centred Gaussian random variables with standard deviation $\frac{\rho}{N^{\alpha}}$. For any $l_1,l_2\in \inti{0}{T}$, we have the following.
	\begin{itemize}
	\item If $l_1+l_2$ is odd, then $B_{T,N}(l_1,l_2)= 0$.
	\item If $l_1+l_2$ is even and $\frac{l_1+l_2}{2}+1\leq N$, then 
	\[\begin{array}{lcl}
		\frac{\rho^{l_1+l_2}}{\frac{l_1+l_2}{2}+1}{l_1+l_2 \choose \frac{l_1+l_2}{2}}
		 \frac{N(N-1)\cdots(N-\frac{l_1+l_2}{2})}{N^{\alpha (l_1+l_2)}} 
		&\leq &
		B_{T,N}(l_1,l_2)\\
		& \leq &
		 \frac{\rho^{l_1+l_2}}{\frac{l_1+l_2}{2}+1}{l_1+l_2 \choose \frac{l_1+l_2}{2}}
		N^{(l_1+l_2)\left(\frac{1}{2}-\alpha\right)+1 } \\
		&&+ \rho^{l_1+l_2}  m_{l_1+l_2}  \left(\frac{l_1+l_2}{2}\right)^{l_1+l_2} N^{(l_1+l_2)\left(\frac{1}{2}-\alpha\right) }.\\
		\end{array}
		\]
	\end{itemize}
	\end{lemma}
	\begin{proof}
	Let $l_1,l_2\in \inti{0}{T}$. Recall the definition
	\[
	B_{T,N}(l_1,l_2)
	=\Ebb \sum\limits_{\substack{1\leq i,i_1, \ldots, i_{l_1}\leq N \\
										1\leq j_1, \ldots, j_{l_2}\leq N}}
		W_{ii_1}W_{i_1i_2}\cdots W_{i_{l_1-1} i_{l_1}}
		W_{ij_{1}}W_{j_{1}j_{2}}\cdots W_{j_{l_2-1}j_{l_2}}.
	\]
	As the odd moments of a centred Gaussian distribution are all null, the above expression shows that indeed $B_{T,N}(l_1,l_2)= 0$ if $l_1+l_2$ is odd. Assume now that $l_1+l_2$ is even. Using the symmetry of $W$, one gets
	\begin{equation}\label{eq:MultiIndexExpectation}
	B_{T,N}(l_1,l_2)
	=\Ebb \sum_{\substack{1\leq i,i_1, \ldots, i_{l_1}\leq N \\
										1\leq j_1, \ldots, j_{l_2}\leq N}}
		 W_{j_{l_2}j_{l_2-1}}\cdots W_{j_2j_1}W_{j_1i}
		W_{ii_{1}}W_{i_{1}i_{2}}\cdots W_{i_{l_1-1}i_{l_1}}
	= \Ebb \sum_{\ibf \in [N]^{l_1+l_2+1}} W_{\ibf}.
	\end{equation}
	
	Let $\ibf=(i_1, i_2, \ldots, i_{l_1+l_2+1}) \in [N]^{l_1+l_2+1}$. Define a relabelling mapping 
	\[g_\ibf: \{i_1, i_2, \ldots, i_{l_1+l_2+1}\}\to \Nbb^*\] 
	by the recursive relation
	\begin{itemize}
		\item $g_\ibf(i_1)=1$,
		\item for $k\in\inti{2}{l_1+l_2+1}$, if there exists $p<k$ such that $i_k=i_p$ then $g_\ibf(i_k)=g_\ibf(i_p)$, otherwise $g_\ibf(i_k)=\max\limits_{1\leq p<k} g_\ibf(i_{p})+1$.
	\end{itemize}
	This way, we see that $\ibf$ defines a unique pair $(G,w)$, where
	\begin{itemize}
		\item $G$ is a connected graph whose vertices are given by the set
	\[
	\{g_\ibf(i_1), g_\ibf(i_2), \ldots, g_\ibf(i_{l_1+l_2+1})\},
	\]
	and whose only edges are the ones connecting $g_\ibf(i_k)$ to $g_\ibf(i_{k+1})$, for $k\in [l_1+l_2]$. Note that $G$ has at most $\min(l_1+l_2+1,N)$ vertices.
		\item $w=(g_\ibf(i_1), g_\ibf(i_2), \ldots, g_\ibf(i_{l_1+l_2+1}))$ is a walk on $G$ of length $l_1+l_2$ crossing all the edges of $G$. Moreover, given any two vertices $m$ and $n$ in $G$, if $m<n$ then $w$ visits the vertex $m$ for the first time before it visits the vertex $n$.
	\end{itemize}
	For instance, if $\ibf=(1,2,3,3,2,5,7)$ then the vertices of $G$ are $\{1,2,3,4,5\}$ and its edges are given by
	\[
	\{1,2\},\{2,3\},\{3\},\{2,4\},\{4,5\},
	\]
	while
	\[
	w=(1,2,3,3,2,4,5).
	\]
	In such cases let us write $\mathbf{i} \sim (G,w)$. Note that this construction is not one-to-one. Given such a pairing $(G,w)$, where $w$ is of length $L$, we can recover all the $(L+1)$-tuples that are mapped to it by injectively mapping the vertices of $G$ to elements in $[N]$. Let us denote the set of all pairings $(G,w)$ obtained in this manner by $\Gcal_{N, l_1+l_2}$.
	
	Next, we note that, for $\ibf \in [N]^{l_1+l_2+1}$, the quantity $\Ebb W_{\ibf}$ depends only on the pair graph-walk $(G,w)$ mapped to it: different $(l_1+l_2+1)$-tuples $\ibf$ associated to the same pair $(G,w)$ yield the same expectation $\Ebb W_{\ibf}$. Let us denote this quantity $E(G,w)$ and remark that 
	\begin{equation}\label{eq:EGwGenericBoundSym}
	0\leq E(G,w)\leq \frac{\rho^{l_1+l_2}}{N^{\alpha (l_1+l_2)}} m_{l_1+l_2}.
	\end{equation}
	
	In summary, we have then
	\begin{equation}\label{eq:WickGraphDecompSym}
	\begin{array}{rcl}
	B_{T,N}(l_1,l_2)
	&=& \sum\limits_{(G,w)\in \Gcal_{N, l_1+l_2}}E(G,w)\cdot 
	\left|\left\{\mathbf{i} \in [N]^{l_1+l_2+1},\;\; \mathbf{i} \sim (G,w) \right\} \right|\\
	&=& \sum\limits_{(G,w)\in \Gcal_{N, l_1+l_2}}E(G,w) N(N-1)\cdots(N-|V_G|+1),\\
	\end{array}
	\end{equation}
	with all the terms in the above sum being positive. For $(G,w)\in \Gcal_{N, l_1+l_2}$, and because all the odd moments of centred Gaussian random variables are equal to $0$, we note that $E(G,w)\neq 0$ if and only if $w$ visits each edge of $G$ an even number of times. We will only consider such pairings in the remainder of this proof. In this case, $w$ is necessarily a loop, $|E_G|\leq \frac{l_1+l_2}{2}$ (as each edge of $G$ is visited at least twice by $w$) and $|V_G|\leq \frac{l_1+l_2}{2}+1$ (by Euler's formula). Let us separate two cases.
	
	\noindent\textbf{Case 1: }$|V_G|=\frac{l_1+l_2}{2}+1$. Note that this case occurs only if $\frac{l_1+l_2}{2}+1\leq N$. This implies in particular that $|E_G|=\frac{l_1+l_2}{2}$ (by Euler's formula). Hence $G$ is a tree and $w$ is a loop on $G$ visiting each edge exactly twice (otherwise we will have the contradiction $|E_G| < \frac{l_1+l_2}{2}$). The numbers of such pairs $(G,w)$ is classically given by (via an identification with Dyck paths, which themselves are related to the Catalan numbers, e.g.\ see \cite{CLRS, HP})
		\[
		\frac{1}{\frac{l_1+l_2}{2}+1}{l_1+l_2 \choose \frac{l_1+l_2}{2}}.
		\]
		For such pairs, $E(G,w)$ can be written as the product of second moments of centred Gaussian random variables and is therefore equal to $\frac{\rho^{l_1+l_2}}{N^{\alpha (l_1+l_2)}}$. By considering only such terms in identity (\ref{eq:WickGraphDecompSym}), we get then the following lower bound
		\[
		\frac{\rho^{l_1+l_2}}{\frac{l_1+l_2}{2}+1}{l_1+l_2 \choose \frac{l_1+l_2}{2}}
		 \frac{N(N-1)\cdots(N-\frac{l_1+l_2}{2})}{N^{\alpha (l_1+l_2)}} \leq 
		B_{T,N}(l_1,l_2).
		\]
		
	\noindent\textbf{Case 2: } $|V_G|\leq \frac{l_1+l_2}{2}$. The number of pairings $(G,w)$ satisfying $|V_G|\leq \frac{l_1+l_2}{2}$ is loosely bounded by $\left(\frac{l_1+l_2}{2}\right)^{l_1+l_2}$ (i.e. the number of ways to construct an arbitrary walk with possible self-loops on a maximum of $\frac{l_1+l_2}{2}$ vertices). Using the computation from the previous case (assuming $\frac{l_1+l_2}{2}+1\leq N$) and the bound (\ref{eq:EGwGenericBoundSym}), we get then
		\[\begin{array}{rcl}
		B_{T,N}(l_1,l_2) &\leq&
		\frac{\rho^{l_1+l_2}}{\frac{l_1+l_2}{2}+1}{l_1+l_2 \choose \frac{l_1+l_2}{2}}
		 \frac{N(N-1)\cdots(N-\frac{l_1+l_2}{2})}{N^{\alpha (l_1+l_2)}} \\ 
		 &&+\sum\limits_{\substack{(G,w)\in \Gcal_{N, l_1+l_2}\\ |V_G|\leq \frac{l_1+l_2}{2}}}E(G,w) N(N-1)\cdots(N-|V_G|+1) \\
		 &\leq&
		\frac{\rho^{l_1+l_2}}{\frac{l_1+l_2}{2}+1}{l_1+l_2 \choose \frac{l_1+l_2}{2}}
		N^{(l_1+l_2)\left(\frac{1}{2}-\alpha\right)+1 }+ 
		\rho^{l_1+l_2}  m_{l_1+l_2}  \left(\frac{l_1+l_2}{2}\right)^{l_1+l_2} N^{(l_1+l_2)\left(\frac{1}{2}-\alpha\right) }.\\ 
		\end{array}
		\]
		
	The proof is now complete.
	\end{proof}
	
	\begin{remark}\label{rem:WsymBHankel} Identity (\ref{eq:MultiIndexExpectation}) shows that $B_{T,N}$ is indeed a Hankel matrix in this (symmetric) case.
	\end{remark}
	
	We are now able to quantify the effect of large reservoir dimension on the separation capacity of a reservoir with symmetric Gaussian connectivity matrix.
	\begin{theorem}\label{Thm:NinftyGaussSym}
	Let $T\in \Nbb^*$, $\rho>0$, $\alpha \in \Rbb$. For $N\in\Nbb^*$, let $W_N\in \Rbb^{N\times N}$ be a symmetric random matrix such that its entries on and above the diagonal are i.i.d.\ centred Gaussian random variables with standard deviation $\frac{\rho}{N^{\alpha}}$. 
	\begin{itemize}
	\item If $\alpha=\frac12$, then 
	\[
	\frac{\lambda_{\max}(B_{T,N})}{N}
	\underset{N\to +\infty}{\longrightarrow} \lambda_{\max}(A_{\mathrm{sc}}(\rho,T))
	\quad\text{and}\quad 
	\frac{\lambda_{\min}(B_{T,N})}{N}
	\underset{N\to +\infty}{\longrightarrow} \lambda_{\min}(A_{\mathrm{sc}}(\rho,T))
	\]
	where $\lambda_{\max}(A_{\mathrm{sc}}(\rho, T))$ (respectively $\lambda_{\min}(A_{\mathrm{sc}}(\rho,T))$) denotes the largest (resp. the smallest) eigenvalue of $A_{\mathrm{sc}}(\rho,T)=((A_{\mathrm{sc}}(\rho,T))_{l_1,l_2})_{0\leq l_1,l_2\leq T}$, the Hankel matrix of moments of order $T$ of the (rescaled) semi-circle law given by
	\[
	(A_{\mathrm{sc}}(\rho,T))_{l_1,l_2}=
	\left\{
	\begin{array}{ll}
	0&\text{if } l_1+l_2 \text{ is odd,}\\
	\frac{\rho^{l_1+l_2}}{\frac{l_1+l_2}{2}+1}{l_1+l_2 \choose \frac{l_1+l_2}{2}}
	&\text{otherwise.}\\
	\end{array}
	\right.
	\]
	Moreover
	\[
	\lim_{N\to \infty}\frac{\lambda_{\max}(B_{T,N})}{\sum\limits_{i=0}^T\lambda_{i} (B_{T,N})} =\frac{\lambda_{\max}(A_{\mathrm{sc}}(\rho,T))}{\mathrm{tr}(A_{\mathrm{sc}}(\rho,T))}.
	\]
	\item If $\alpha<\frac12$, then 
	\[
	\lambda_{\max}(B_{T,N}) 
	\underset{N\to +\infty}{\simeq} 
	\frac{\rho^{2T}}{T+1}{2T \choose T} N^{1+2T\left(\frac{1}{2}-\alpha\right)},
	\]
	and $\lambda_{\max}(B_{T,N})$ dominates the spectrum of $B_{T,N}$ as $N$ goes to infinity, i.e.
	\[
	\lim_{N\to \infty}\frac{\lambda_{\max}(B_{T,N})}{\sum\limits_{i=0}^T\lambda_{i} (B_{T,N})} =1.
	\]
	\item If $\alpha>\frac12$, then 
	\[
	\lambda_{\max}(B_{T,N}) 
	\underset{N\to +\infty}{\simeq} 
	N,
	\]
	and $\lambda_{\max}(B_{T,N})$ dominates the spectrum of $B_{T,N}$ as $N$ goes to infinity.
	\end{itemize}
	\end{theorem}
	\begin{proof}
	For large $N$, Lemma \ref{lemma:OrderMagnitudeSymmetric} implies 
	\[
	\begin{array}{ll}
		\sum\limits_{t=0}^T \frac{\rho^{2t}}{t+1}{2t \choose t} \frac{N(N-1)\cdots(N-t)}{N^{2 \alpha t}}  &\leq 
		\sum\limits_{l=0}^T B_{T,N}(l,l)\\
		& \leq 
		\sum\limits_{t=0}^T  \left( \rho^{2t}m_{2t} t^{2t} N^{2t\left(\frac{1}{2}-\alpha\right)} +
		  \frac{\rho^{2t}}{t+1}{2t \choose t} N^{1+2t\left(\frac{1}{2}-\alpha\right)} \right).\\
		\end{array}
	\]
	Hence
	\[
	\mathrm{tr}(B_{T,N}) 
	\underset{N\to +\infty}{\simeq}  
	\sum_{t=0}^T\frac{\rho^{2t}}{t+1}{2t \choose t} N^{1+2t\left(\frac{1}{2}-\alpha\right)}.
	\]
	Similarly
	\[
	\|B_{T,N}\|_{2,\infty}\underset{N\to +\infty}{\simeq}  
	\max_{0\leq l_1 \leq T} \sqrt{
	\sum\limits_{\substack{0\leq l_2 \leq T \\ l_1 + l_2 \text{ even}} }  \left( 
		\frac{\rho^{l_1+l_2}}{\frac{l_1+l_2}{2}+1}{l_1+l_2 \choose \frac{l_1+l_2}{2}}
		N^{(l_1+l_2)\left(\frac{1}{2}-\alpha\right)+1 }
		\right)^2
	}.
	\]
	
	We have the following cases.\\
	\textbf{Case 1: } $\alpha<\frac12$. In this case
	\[\mathrm{tr}(B_{T,N})\underset{N\to +\infty}{\simeq} B_{T,N}(T,T) \underset{N\to +\infty}{\simeq} \|B_{T,N}\|_{2,\infty}.\]
	Therefore, following Proposition \ref{prop:EVboundsNdim}
	\[
	\lambda_{\max}(B_{T,N}) 
	\underset{N\to +\infty}{\simeq} 
	\frac{\rho^{2T}}{T+1}{2T \choose T} N^{1+2T\left(\frac{1}{2}-\alpha\right)},
	\]
	and $\lambda_{\max}(B_{T,N})$ dominates the spectrum of $B_{T,N}$ as $N$ goes to infinity
	\[
	\lim_{N\to \infty}\frac{\lambda_{\max}(B_{T,N})}{\sum\limits_{i=0}^T\lambda_{i} (B_{T,N})} =
	\lim_{N\to \infty}\frac{\lambda_{\max}(B_{T,N})}{\mathrm{tr}(B_{T,N})} =
	1.
	\]
	\noindent \textbf{Case 2: }$\alpha>\frac12$. Similarly
	\[
	\lambda_{\max}(B_{T,N}) 
	\underset{N\to +\infty}{\simeq} 
	B_{T,N}(0,0) \underset{N\to +\infty}{\simeq} 
	N,
	\]
	and $\lambda_{\max}(B_{T,N})$ dominates the spectrum of $B_{T,N}$ as $N$ goes to infinity.\\
	\textbf{Case 3: } $\alpha=\frac12$. By Lemma \ref{lemma:OrderMagnitudeSymmetric}, for all $l_1,l_2 \in \inti{0}{T}$
	\[
	\frac{B_{T,N}(l_1,l_2)}{N}
	\underset{N\to +\infty}{\longrightarrow} 
	(A_{\mathrm{sc}}(\rho,T))_{l_1,l_2}=
	\left\{
	\begin{array}{ll}
	0&\text{if } l_1+l_2 \text{ is odd,}\\
	\frac{\rho^{l_1+l_2}}{\frac{l_1+l_2}{2}+1}{l_1+l_2 \choose \frac{l_1+l_2}{2}}
	&\text{otherwise.}\\
	\end{array}
	\right.
	\]
	The claim follows trivially from the above.
	\end{proof}
	\begin{remark}
	$A_{\mathrm{sc}}(\rho, T)$ corresponds to the Hankel matrix of moments (as defined in Remark \ref{rem:HankelDim1}) of the rescaled Wigner semi-circle law whose density is given by
	\[
	\frac{1}{2\pi \rho^2}\sqrt{\max(4\rho^2 - x^2, 0)}
	,\;\;\text{for all } x\in \Rbb.
	\]
	\end{remark}
	\begin{remark}\label{rem:RescaleUSym} The extra $N$ factor appearing in Theorem \ref{Thm:NinftyGaussSym} (for example in the limit case for $\alpha=\frac12$) is merely due to the norm of the chosen pre-processing matrix $u$. This factor will disappear if we choose $u$ to be, for instance, the unit vector $\frac{1}{\sqrt{N}}(1,\ldots, 1)^\top$.
	\end{remark}
	
	Theorem~\ref{Thm:NinftyGaussSym} highlights a sharp distinction between the classical scaling $\alpha=\frac12$ and all other normalisations. When the entries of $W$ are properly rescaled, i.e. $\alpha=\frac12$, the spectrum of $B_{T,N}$ admits a non-trivial limiting distribution governed by the moment structure of the Wigner semi-circle law. In particular, several eigendirections (in the space of time series $\Rbb^{T+1}$) contribute meaningfully to the expected separation of time series. By contrast, when $\alpha\neq\frac12$, the expected separation becomes effectively one-dimensional: the geometry induced by the reservoir is governed by a single dominant eigendirection.
	
\begin{figure}[h!]
  \centering
  \begin{subfigure}{0.32\textwidth}
    \centering
    \adjustbox{valign=c}{\includegraphics[scale=0.33]{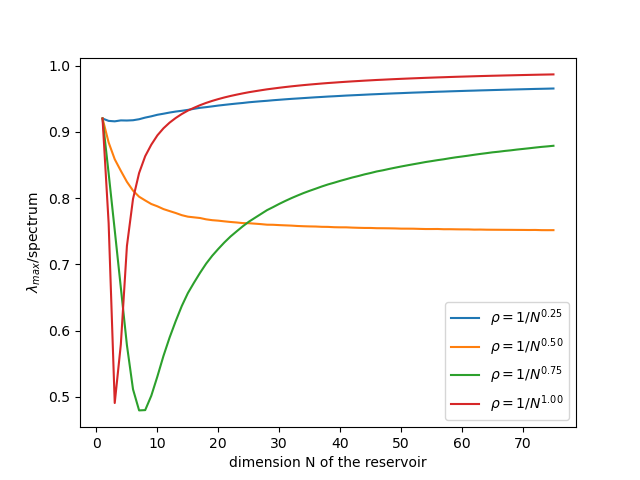}}
    \caption{$T=6$}
  \end{subfigure}
  \begin{subfigure}{0.32\textwidth}
    \centering
    \adjustbox{valign=c}{\includegraphics[scale=0.33]{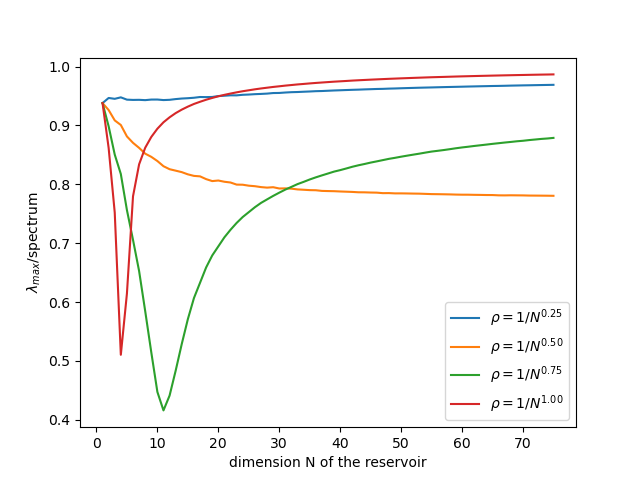}}
    \caption{$T=10$}
  \end{subfigure}
  \begin{subfigure}{0.32\textwidth}
    \centering
    \adjustbox{valign=c}{\includegraphics[scale=0.33]{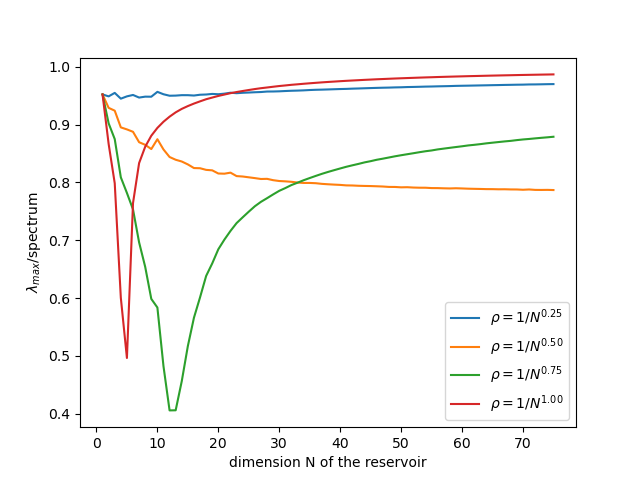}}
    \caption{$T=12$}
  \end{subfigure}
  \caption{The evolutions in function of the dimension $N$ of the reservoir of the dominance ratio $r_{T,N}$ (as defined in (\ref{eq:DomRatioN})) of the generalised matrix of moments associated to a symmetric $N\times N$ random connectivity matrix with i.i.d.\ entries on and above the diagonal, which are centred Gaussians with standard deviation $\rho=\frac{1}{N^{\alpha}}$. The different panels correspond to different lengths $T$ of the times-series, while the different curves correspond to different values of the scaling exponent $\alpha$.}
  \label{Fig:GaussEVNNinfty}
	\end{figure}

	Figure~\ref{Fig:GaussEVNNinfty}
	\footnote{The code to replicate all the figures and tables in this paper is publicly available on \href{https://github.com/youness-boutaib/linear-separation-capacity}{https://github.com/youness-boutaib/linear-separation-capacity}.}
	 illustrates this phenomenon through the evolution of the dominance ratio $r_{T,N}$ (as defined in (\ref{eq:DomRatioN})) as a function of the reservoir dimension $N$ for different values of $T$. We simulated the matrices $B_{T,N}$ using a Monte Carlo method and computed their eigenvalues using the `eigvalsh' function \footnote{\href{https://numpy.org/doc/stable/reference/generated/numpy.linalg.eigvalsh.html}{https://numpy.org/doc/stable/reference/generated/numpy.linalg.eigvalsh.html}} in Numpy. Note that the combination of computing large powers of large matrices and using a Monte Carlo simulation results in the loss of numerical accuracy and stability in the computed eigenvalues. This is why we limit ourselves here to the rather modest value $T=12$.

\begin{figure}[h!]
	\includegraphics[scale=0.4]{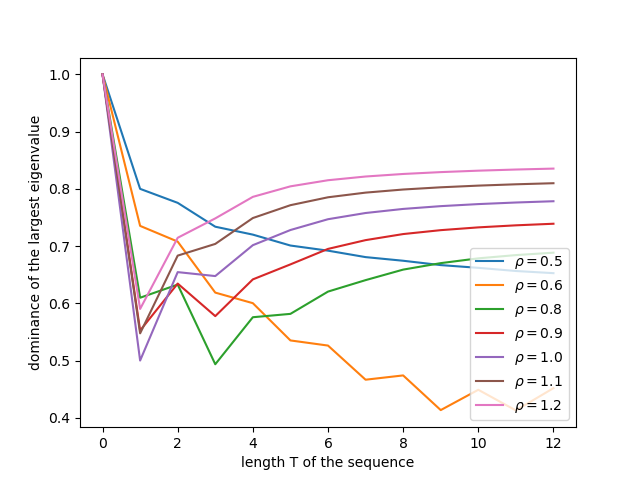}
	\centering
	\caption{The dominance $r_{T}$ of the largest eigenvalue, in function of the length $T$, over the entire spectrum of the Hankel matrix of moments $A_{\mathrm{sc}}(\rho,\cdot)$ of the rescaled Wigner semi-circle law with rescaling parameter $\rho$.}
	\label{Fig:Lambda_dominance_dim_1_WignerSC}
	\end{figure}

	To visualise the limiting regime ($N\to\infty$) associated with the critical scaling $\alpha=\frac12$, Figure~\ref{Fig:Lambda_dominance_dim_1_WignerSC} displays the asymptotic dominance ratio 
	\[r_T 
	= \frac{\lambda_{\max}(A_{\mathrm{sc}}(\rho,T))}{\mathrm{tr}(A_{\mathrm{sc}}(\rho,T))}.
	\]
	A striking feature is that, over a substantial range of values of $T$ (depending on $\rho$), the dominance ratio $r_T$ remains well below $1$. Hence the limiting geometry induced by the semi-circle law retains a genuinely multi-dimensional character. Given the universality of the limiting Wigner semi-circle law (e.g.\ \cite[Chapter~2]{AGZ}), one may reasonably expect this result to generalise for broad classes of symmetric random reservoirs.

	We turn our attention now to the study of the separation capacity of the reservoir for very long time series. For this purpose, we study the entries of $B_{T,N}$ in function of $T$.
	\begin{lemma}\label{lemma:OrderMagnitudeSymmetricTinf}
	Let $N\in \Nbb^*$ and $\sigma>0$. Let $W\in \Rbb^{N\times N}$ be a symmetric random matrix such that its entries on and above the diagonal are i.i.d.\ centred Gaussian random variables with standard deviation $\sigma$. For $T\in \Nbb^*$ and all $l_1,l_2\in \inti{0}{T-1}$, one has
	\[
	B_{T,N}(l_1,l_2)
	\leq 
	\frac{1}{\sigma^2} \frac{1}{\frac{l_1+l_2}{\frac{N(N+1)}{2} }+1} B_{T,N}(l_1+1,l_2+1).
	\]
	\end{lemma}
	\begin{proof} The inequality is trivial if $l_1+l_2$ is odd. Assume that $l_1+l_2$ is even. Let us recall the following identity obtained in the proof of Lemma \ref{lemma:OrderMagnitudeSymmetric}
	\[
	B_{T,N}(l_1,l_2)
	= \sum\limits_{(G,w)\in \Gcal_{N, l_1+l_2}}E(G,w) N(N-1)\cdots(N-|V_G|+1),
	\]
	with $w$ being a loop visiting each edge of $G$ an even number of times. For such a loop $w$, let $k_w=2p_w$ be the largest amount of times an edge of $G$ is visited by $w$. Note that, trivially 
	\[\frac{l_1+l_2}{\frac{N(N+1)}{2} }\leq
	\frac{l_1+l_2}{|E_G|}\leq k_w\leq l_1+l_2.\]
	Let $\tilde{e}(w)$ be the first edge of $G$ that is crossed $k_w$ times by $w$. Let $\tilde{w}$ be the walk on $G$ obtained from $w$ by adding a double-crossing of $\tilde{e}(w)$ after it is crossed $k_w$ times by $w$. For example, if $w=(1,2,3,3,3,2,1)$, then all edges are visited twice and $k_w=2$. Since the self-loop $\{3\}$ is the first edge visited $k_w$ times by $w$, then
	\[
	\tilde{w} = (1,2,3,3,3,3,3,2,1).
	\] 
	Note that the mapping 
	\[
	(G,w)\in \Gcal_{N, l_1+l_2} \mapsto (G,\tilde{w})\in \Gcal_{N, l_1+l_2+2}
	\]
	is injective, but not surjective. With the introduced notation, we have the bound
		\[
		E(G,w) = \frac{1}{\sigma^2} \frac{m_{k_w}}{m_{k_w+2}}E(G,\tilde{w})
		=\frac{1}{\sigma^2} \frac{1}{k_w+1}E(G,\tilde{w})
		\leq \frac{1}{\sigma^2} \frac{1}{\frac{l_1+l_2}{\frac{N(N+1)}{2} }+1}E(G,\tilde{w}).
		\]
	Hence, we get
	\[\begin{array}{rcl}
	B_{T,N}(l_1,l_2)
	&\leq & 
	\frac{1}{\sigma^2}\frac{1}{\frac{l_1+l_2}{\frac{N(N+1)}{2} }+1}
	\sum\limits_{(G,w)\in \Gcal_{N, l_1+l_2}}E(G,\tilde{w}) N(N-1)\cdots(N-|V_G|+1)\\
	&\leq & 
	\frac{1}{\sigma^2}\frac{1}{\frac{l_1+l_2}{\frac{N(N+1)}{2} }+1}
	\sum\limits_{(G,w)\in \Gcal_{N, l_1+l_2+2}}E(G, w) N(N-1)\cdots(N-|V_G|+1)\\
	&=& \frac{1}{\sigma^2}\frac{1}{\frac{l_1+l_2}{\frac{N(N+1)}{2} }+1} B_{T,N}(l_1+1,l_2+1).\\
	\end{array}
	\]
	\end{proof}
	
	The next theorem formalises a rather naturally expected result: that the quality of separation deteriorates with time, irrespective of the chosen scaling.
	\begin{theorem}\label{Thm:TinftyGaussSym} Let $N\in \Nbb^*$. Let $W\in \Rbb^{N\times N}$ be a symmetric random matrix such that its entries on and above the diagonal are i.i.d.\ centred Gaussian random variables. 
	Then, as $T\to \infty$, $\lambda_{\mathrm{max}} (B_{T,N})$ dominates the spectrum of $B_{T,N}$, i.e.
	\[
	\frac{\lambda_{\mathrm{max}} (B_{T,N})}{\sum\limits_{i=0}^{T}\lambda_i(B_{T,N})}
	\underset{T\to \infty}{\longrightarrow} 1.
	\]
	Moreover $\lambda_{\mathrm{max}} (B_{T,N})\underset{T\to +\infty}{\simeq} B_{T,N}(T,T)$.
	\end{theorem}
	\begin{proof}
	Denote by $\sigma$ the standard deviation of the entries of $W$. Following Lemma \ref{lemma:OrderMagnitudeSymmetricTinf}, one has for all $l\in \inti{0}{T-1}$
	\[
	B_{T,N}(l,l)
	\leq \frac{1}{\sigma^{2(T-l)}}
	 \prod_{t=l}^{T-1}  \frac{1}{\frac{2t}{\frac{N(N+1)}{2} }+1}
	 B_{T,N}(T,T).
	\]
	Therefore
	\[
	1 \leq 
	\frac{\mathrm{tr}(B_{T,N})}{B_{T,N}(T,T)}
	\leq 
	1+ \sum_{l=0}^{T-1} \frac{1}{\sigma^{2(T-l)}}
	 \prod_{t=l}^{T-1}  \frac{1}{\frac{2t}{\frac{N(N+1)}{2} }+1}.
	\]
	For any two strictly positive numbers $a$ and $b$, the sequence $(p_t)_{t \geq 0}$ given by 
	\[
	p_0=1
	\quad \text{and}\quad
	p_t = \frac{1\cdot (b+1)\cdot (2b+1) \cdots ((t-1)b+1)}{a^t}\quad\text{for all }t\geq 1.
	\]
	diverges to $+\infty$. Moreover, it is strictly increasing from a certain rank (that depends on $a$ and $b$.) Hence, there exists $t_0\geq 1$ (depending on $a$ and $b$) such that $p_t\leq p_{t_0}$ for all $1\leq t\leq t_0$, and $p_t\geq p_{s}$ for all $t\geq s \geq t_0$. In particular, one has, for all $T\geq t_0+2$
	\[
	\sum_{l=0}^{T-2} p_l
	\leq (T-1)p_{T-2}.
	\]
	Applying this in the case where $a=\frac{1}{\sigma^{2}}$ and $b=\frac{4}{N(N+1)}$, we get that
	\[
	\sum_{l=0}^{T-2} \frac{p_l}{p_T}
	=\sum_{l=0}^{T-2} \frac{1}{\sigma^{2(T-l)}}
	 \prod_{t=l}^{T-1}  \frac{1}{\frac{2t}{\frac{N(N+1)}{2} }+1}
	\leq (T-1) \frac{p_{T-2}}{p_T}
	 = \frac{1}{\sigma^{4}}
	 \frac{(T-1)}{\left(\frac{4(T-1)}{N(N+1)}+1\right)\left(\frac{4(T-2)}{N(N+1)}+1\right)}.
	\]
	Hence
	\[
	1 \leq 
	\frac{\mathrm{tr}(B_{T,N})}{B_{T,N}(T,T)}
	\leq 
	1+ \frac{1}{\sigma^{2}}
	 \frac{1}{\left(\frac{4(T-1)}{N(N+1)}+1\right)} 
	 + \frac{1}{\sigma^{4}}
	 \frac{(T-1)}{\left(\frac{4(T-1)}{N(N+1)}+1\right)\left(\frac{4(T-2)}{N(N+1)}+1\right)}.
	\]
	and $ \lim\limits_{T\to\infty} \frac{\mathrm{tr}(B_{T,N})}{B_{T,N}(T,T)}= 1$. As a direct consequence of Proposition \ref{prop:EVboundsNdim}, one has, for all $T\in \Nbb$
	\[
	1\leq 
	\frac{\lambda_{\mathrm{max}} (B_{T,N}) }{B_{T,N}(T,T)}
	\leq
	\frac{\mathrm{tr}(B_{T,N})}{B_{T,N}(T,T)}.
	\]
	Therefore, one also has $ \lim\limits_{T\to\infty} \frac{\lambda_{\mathrm{max}} (B_{T,N})}{B_{T,N}(T,T)}= 1$. Thus
	\[
	\frac{\lambda_{\mathrm{max}} (B_{T,N})}{\sum\limits_{t=0}^{T}\lambda_i(B_{T,N})}= 
	\frac{\lambda_{\mathrm{max}} (B_{T,N})}{\mathrm{tr}(B_{T,N})} 
	\underset{T\to \infty}{\longrightarrow} 1.
	\]
	\end{proof}
	
\begin{figure}[h!]
  \centering
  \begin{subfigure}{0.32\textwidth}
    \centering
    \adjustbox{valign=c}{\includegraphics[scale=0.33]{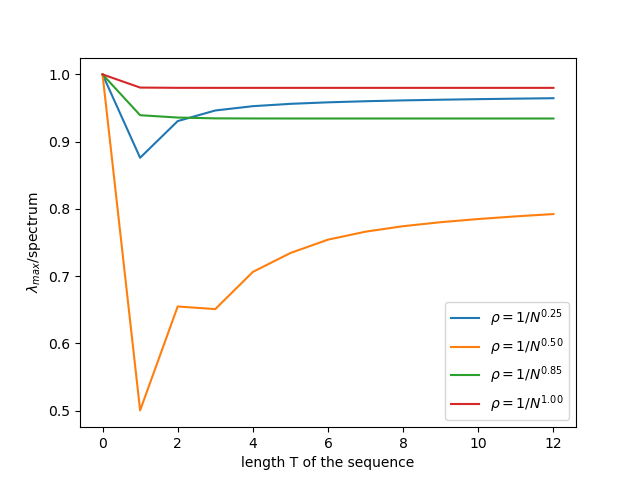}}
    \caption{$N=50$}
  \end{subfigure}
  \begin{subfigure}{0.32\textwidth}
    \centering
    \adjustbox{valign=c}{\includegraphics[scale=0.33]{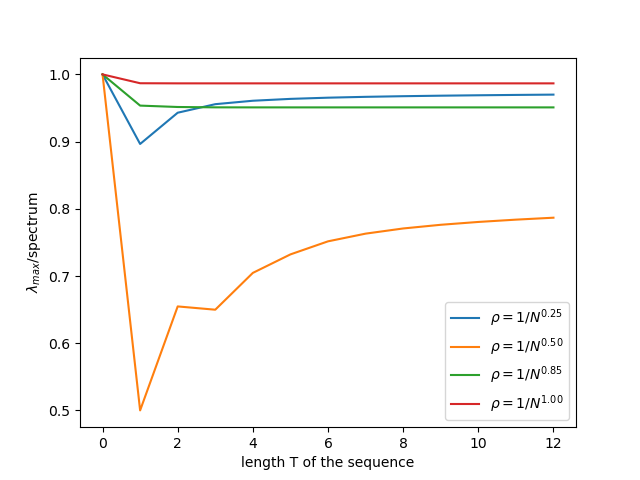}}
    \caption{$N=75$}
  \end{subfigure}
  \begin{subfigure}{0.32\textwidth}
    \centering
    \adjustbox{valign=c}{\includegraphics[scale=0.33]{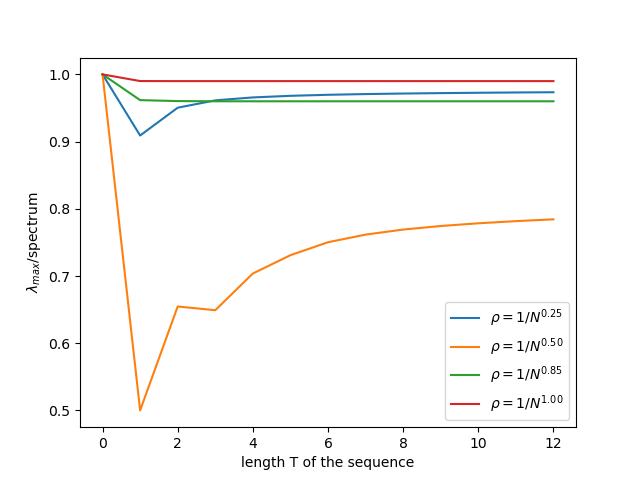}}
    \caption{$N=100$}
  \end{subfigure}
  \caption{The evolutions in function of the length of the time series $T$ of the dominance ratio $r_{T,N}$ (as defined in (\ref{eq:DomRatioN})) of the generalised matrix of moments associated to a symmetric $N\times N$ random connectivity matrix with i.i.d.\ entries on and above the diagonal, which are centred Gaussians with standard deviation $\rho=\frac{1}{N^{\alpha}}$. The different panels correspond to different dimensions $N$ of the reservoir, while the different curves correspond to different values of the scaling exponent $\alpha$.}
  \label{Fig:GaussEVNTinftySym}
  \end{figure}

	Figure \ref{Fig:GaussEVNTinftySym} illustrates the convergence of the dominance ratio $r_{T,N}$ as $T\to \infty$ for different values of the reservoir dimension $N$.

  Although Figure \ref{Fig:GaussEVNTinftySym} may suggest at least a much slower convergence towards $1$ of the dominance ratio $r_{T,N}$ in the case of a scaling $N^{-\frac12}$, Theorem \ref{Thm:TinftyGaussSym} proves that $r_{T,N}$ converges eventually to $1$. Combining Theorems~\ref{Thm:NinftyGaussSym} and~\ref{Thm:TinftyGaussSym}, we arrive at the following picture. For symmetric Gaussian reservoirs, the classical scaling $N^{-\frac12}$ is the unique normalisation that preserves a non-trivial spectral geometry when the reservoir dimension grows. Nevertheless, for sufficiently long time series, the largest eigenvalue eventually dominates the entire spectrum, implying an unavoidable loss of diversity in the directions along which signals are separated. This suggests that symmetric random reservoirs are naturally better suited to short and moderate time horizons, while large reservoir dimensions combined with the critical scaling $N^{-\frac12}$ provide the most favourable separation properties within this class of models.
	\subsection{The i.i.d.\ case}\label{subsec:ExpIID}
\begin{figure}[h!]
  \centering
  \begin{subfigure}{0.32\textwidth}
    \centering
    \adjustbox{valign=c}{\includegraphics[scale=0.33]{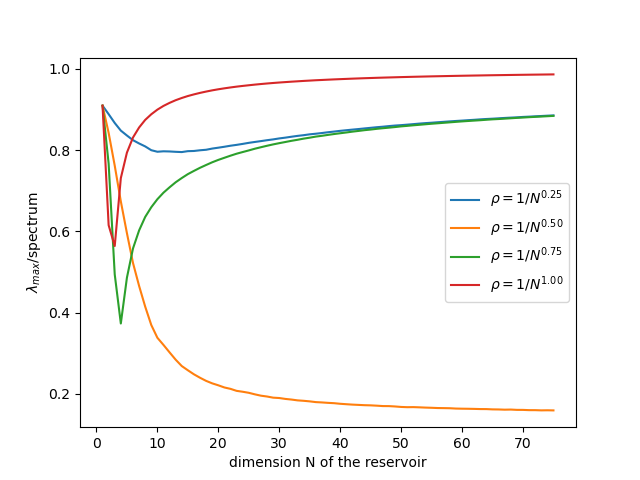}}
    \caption{$T=6$}
  \end{subfigure}
  \begin{subfigure}{0.32\textwidth}
    \centering
    \adjustbox{valign=c}{\includegraphics[scale=0.33]{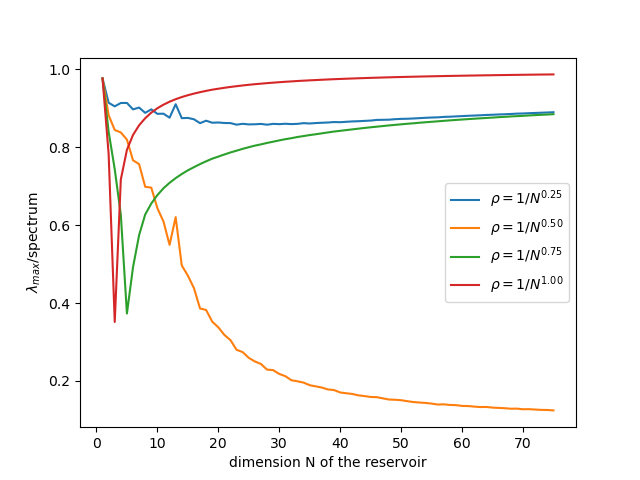}}
    \caption{$T=10$}
  \end{subfigure}
  \begin{subfigure}{0.32\textwidth}
    \centering
    \adjustbox{valign=c}{\includegraphics[scale=0.33]{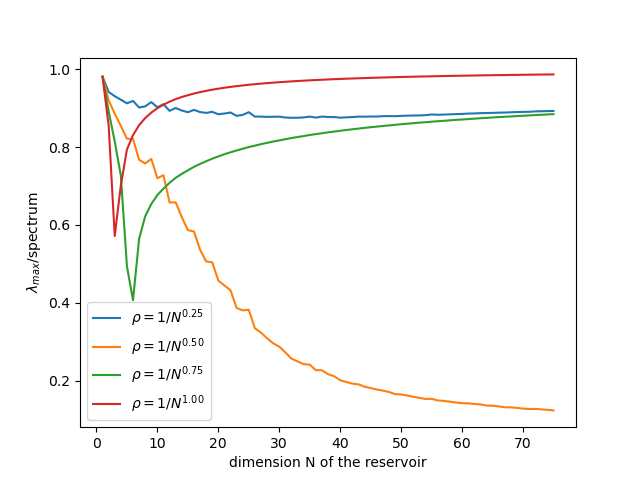}}
    \caption{$T=12$}
  \end{subfigure}
  \caption{The evolutions in function of the dimension $N$ of the reservoir of the dominance ratio $r_{T,N}$ (as defined in (\ref{eq:DomRatioN})) of the generalised matrix of moments associated to an $N\times N$ random connectivity matrix with i.i.d.\ centred Gaussian entries with standard deviation $\rho=\frac{1}{N^{\alpha}}$. The different panels correspond to different lengths $T$ of the times-series, while the different curves correspond to different values of the scaling exponent $\alpha$.}
  \label{Fig:GaussEVNNinftyIID}
	\end{figure}

	We now turn to the case in which the entries of the connectivity matrix are independent and identically distributed. In contrast with the symmetric setting, considerably fewer results are available in the random matrix literature regarding the computation of moments of the type appearing in Definition \ref{defn:GenHankelMatrix}. One reason is that these moments are not directly related to the convergence of the empirical spectral distribution towards the circular law (\cite{Bai,TV}), in the same way that the classical moment method is connected to the Wigner semi-circle law in the symmetric case. At the time of writing, we were only aware of \cite{AD} leading similar computations. Nevertheless, the combinatorial arguments remain very close in spirit to those of the previous subsection. The main difference is that one must now consider the direction of the walks on the graphs. As before, let us first introduce two pieces of notation that will be used throughout this section.
	\begin{notation}
	Let $p,q, N\in \Nbb^*$. Given $\ibf=(i_1, i_2, \ldots, i_p) \in [N]^{p}$ and $\jbf=(i_p, j_2, \ldots, j_q) \in [N]^{q}$, we denote by $\ibf\jbf$ their concatenation given as the $(p+q-1)$-tuple
	\[
	\ibf\jbf:=(i_1, i_2, \ldots, i_p, j_2, \ldots, j_q).
	\]
	We denote by $\bar{\ibf}$ the reverse $p$-tuple
	\[
	\bar{\ibf} := (i_p, i_{p-1}, \ldots, i_1).
	\]
	\end{notation}
	Let us now move to our main estimate.
	\begin{lemma}\label{lemma:OrderMagnitudeIID}
	Let $N, T\in \Nbb^*$, $\rho>0$ and $\alpha \in \Rbb$. Let $W\in \Rbb^{N\times N}$ be a random matrix such that its entries are i.i.d.\ centred Gaussian random variables with standard deviation $\frac{\rho}{N^{\alpha}}$. For $l_1,l_2\in \inti{0}{T}$, we have the following.
	\begin{itemize}
	\item If $l_1+l_2$ is odd, then $B_{T,N}(l_1,l_2)= 0$,
	\item If $l_1+l_2$ is even and $\frac{l_1+l_2}{2}+1\leq N$, then 
	\[\begin{array}{lcl}
		\rho^{l_1+l_2} \frac{N(N-1)\cdots(N-\frac{l_1+l_2}{2})}{N^{\alpha (l_1+l_2)}}  \delta_{l_1,l_2}
		&\leq &
		B_{T,N}(l_1,l_2)\\
		& \leq &
		\rho^{l_1+l_2} N^{(l_1+l_2)\left(\frac{1}{2}-\alpha\right)+1 } \delta_{l_1,l_2}\\
		&&+ \rho^{l_1+l_2}  m_{l_1+l_2}  \left(\frac{l_1+l_2}{2}\right)^{l_1+l_2} N^{(l_1+l_2)\left(\frac{1}{2}-\alpha\right) },\\
		\end{array}
		\]
		where $\delta_{l_1,l_2}$ is the Kronecker delta symbol.
	\end{itemize}
	\end{lemma}
	\begin{proof}
	Let $l_1,l_2\in \inti{0}{T}$. As in the proof of Lemma \ref{lemma:OrderMagnitudeSymmetric}, $B_{T,N}(l_1,l_2)$ is trivially null if $l_1+l_2$ is odd. We assume then that $l_1+l_2$ is even. Expanding $B_{T,N}(l_1,l_2)$, one gets
	\[
	B_{T,N}(l_1,l_2)
	=\Ebb \sum\limits_{\substack{1\leq i,i_1, \ldots, i_{l_1}\leq N \\
										1\leq j_1, \ldots, j_{l_2}\leq N}}
		W_{ii_1}W_{i_1i_2}\cdots W_{i_{l_1-1} i_{l_1}}
		W_{ij_{1}}W_{j_{1}j_{2}}\cdots W_{j_{l_2-1}j_{l_2}}.
	\]
	In the above expression, if $j_{l_2}\neq i_{l_1}$, then the tuple
	\[
	(j_{l_2}, j_{l_2-1}, \ldots, j_{1}, i, i_1, i_2, \ldots, i_{l_1})
	\]
	describes a walk on a graph (whose vertices are given by $[N]$) that does not end where it starts. Therefore, one of the edges of this graph is crossed (in some direction) an odd number of times, which implies that the corresponding expectation is null
	\[
	\Ebb W_{ii_1}W_{i_1i_2}\cdots W_{i_{l_1-1} i_{l_1}}
		W_{ij_{1}}W_{j_{1}j_{2}}\cdots W_{j_{l_2-1}j_{l_2}} = 0.
	\]
	Consequently
		\begin{equation}\label{eq:WkWiWj}
		\begin{array}{rcl}
		B_{T,N}(l_1,l_2) 
	&=& \sum\limits_{\substack{1\leq i,i_1, \ldots, i_{l_1-1}\leq N \\
										1\leq j, j_1, \ldots, j_{l_2-1}\leq N}}
		W_{ii_1}W_{i_1i_2}\cdots W_{i_{l_1-1} j}
		W_{ij_{1}}W_{j_{1}j_{2}}\cdots W_{j_{l_2-1}j}\\
	&=& \sum\limits_{\kbf=\ibf\overline{\jbf}\in \Ical_2(l_1+l_2+1)}
		\Ebb W_{\ibf}W_{\jbf},\\
		\end{array}
	\end{equation}
	where, in the above formula,
	\begin{itemize}
		\item we implicitly assume in the decomposition $\kbf=\ibf\overline{\jbf}$ that $\ibf=(i,i_1, \ldots, i_{l_1-1},j)$ has $l_1+1$ elements and that $\jbf=(i,j_1, \ldots, j_{l_2-1},j)$ has $l_2+1$ elements (and both start and end with the same elements.)
		\item $\Ical_2(l_1+l_2+1)$ denotes the set of tuples $\kbf=(k_1, k_2, \ldots, k_{l_1+l_2+1})$ where the first and last elements are equal ($k_1= k_{l_1+l_2+1}$), and where, for all $i\in [l_1+l_2]$, there exists an even number of elements $p\in [l_1+l_2]$ such that $\{k_i,k_{i+1}\}=\{k_p,k_{p+1}\}$ (otherwise, we will have $\Ebb W_{\ibf}W_{\jbf}=0$).
	\end{itemize}
	
	Exactly as in Lemma \ref{lemma:OrderMagnitudeSymmetric}, we map each tuple $\kbf$ to a pair $(G,w)$, with $G$ being a graph and $w$ a walk on $G$. In such cases let us denote again $\kbf \sim (G,w)$. Given such a pairing $(G,w)$, where $w$ is of length $L$, we can recover all the $(L+1)$-tuples $\kbf$ that are mapped to it by injectively mapping the vertices of $G$ to elements in $[N]$. Since we are restricting ourselves in identity (\ref{eq:WkWiWj}) to tuples starting and ending at the same points, then the corresponding walk $w$ is necessarily a loop. Let us denote the set of all pairings $(G,w)$ obtained in this manner by $\Lcal_{N, l_1+l_2}$. As we are restricting ourselves to the set of indices $\Ical_2(l_1+l_2+1)$ (where each edge is visited an even number of times), then $|E_G|\leq \frac{l_1+l_2}{2}$ and $|V_G|\leq \frac{l_1+l_2}{2}+1$ (by Euler's formula).
	
	We note again, that given $\kbf=\ibf\overline{\jbf} \in [N]^{l_1+l_2+1}$, the quantity $\Ebb W_{\ibf}W_{\jbf}$ depends only on the pair graph-walk $(G,w)$ associated with $\kbf$. We denote this quantity $E_{l_1}(G,w)$ and remark again that 
	\begin{equation}\label{eq:EGwGenericBoundIID}
	0\leq E_{l_1}(G,w)\leq \frac{\rho^{l_1+l_2}}{N^{\alpha (l_1+l_2)}} m_{l_1+l_2}.
	\end{equation}
	In light of this, Identity \eqref{eq:WkWiWj} becomes
	\[\begin{array}{rcl}
	B_{T,N}(l_1,l_2) 
	&=& \sum\limits_{(G,w)\in \Lcal_{N, l_1+l_2}}E_{l_1}(G,w)\cdot 
	\left|\left\{\kbf \in \Ical_2(l_1+l_2+1),\;\; \kbf \sim (G,w) \right\} \right|\\
	&=&  \sum\limits_{(G,w)\in \Lcal_{N, l_1+l_2}}E_{l_1}(G,w) N(N-1)\cdots(N-|V_G|+1) .\\
	\end{array}
	\]
	In the above sum, we will separate again two cases.
	
	\noindent\textbf{Case 1: } $|V_G|=\frac{l_1+l_2}{2}+1$ (assuming $\frac{l_1+l_2}{2}+1\leq N$). Then, by Euler's formula, $|E_G|=\frac{l_1+l_2}{2}$ and $G$ is a tree. Consequently, $w$ crosses each edge of $G$ exactly twice. More exactly, $w$ crosses each edge of $G$ exactly once in each direction (otherwise $G$ will contain a cycle, which will contradict that it is a tree). Let $\kbf=\ibf\overline{\jbf}\sim (G,w)$ be a tuple satisfying this condition and decompose $w=w_1 \overline{w_2}$, where $w_1$ corresponds to $\ibf$ and $w_2$ corresponds to $\jbf$. If an edge is crossed twice (in different directions as per our assumptions) in $w_i$, for $i\in \{1,2\}$, then, as this edge cannot be crossed another time in $w$, we get  $E_{l_1}(G,w)=0$. Hence, we are left with a single configuration, where each edge is crossed once in $w_1$ (corresponding to $\ibf$) then crossed another time in the reverse direction in $\overline{w_2}$ or, equivalently, crossed in the same direction in $w_2$ (corresponding to $\jbf$). Obviously, this can only happen if $l_1=l_2$. In this case, $E_{l_1}(G,w)$ is a product of second moments of independent Gaussian random variables. More precisely, we have
		\[
		E_{l_1}(G,w) = \frac{\rho^{l_1+l_2}}{N^{\alpha (l_1+l_2)}}.
		\]
		This implies that
		\[
		\rho^{l_1+l_2} \frac{N(N-1)\cdots(N-\frac{l_1+l_2}{2})}{N^{\alpha (l_1+l_2)}}  \delta_{l_1,l_2}
		\leq 
		B_{T,N}(l_1,l_2).
		\]
		
	\noindent\textbf{Case 2:} $|V_G|\leq \frac{l_1+l_2}{2}$. The number of pairings $(G,w)$ satisfying $|V_G|\leq \frac{l_1+l_2}{2}$ is loosely bounded by $\left(\frac{l_1+l_2}{2}\right)^{l_1+l_2}$. Using the computation from the previous case when $\frac{l_1+l_2}{2}+1\leq N$ and the bound (\ref{eq:EGwGenericBoundIID}), we get
		\[\begin{array}{rcl}
		B_{T,N}(l_1,l_2) &\leq&
		\rho^{l_1+l_2}
		 \frac{N(N-1)\cdots(N-\frac{l_1+l_2}{2})}{N^{\alpha (l_1+l_2)}} \delta_{l_1,l_2} \\
		 &&+
		 \sum\limits_{\substack{(G,w)\in \Lcal_{N, l_1+l_2}\\ |V_G|\leq \frac{l_1+l_2}{2}}}E_{l_1}(G,w) N(N-1)\cdots(N-|V_G|+1) \\ 
		 &\leq&
		\rho^{l_1+l_2} N^{(l_1+l_2)\left(\frac{1}{2}-\alpha\right)+1 } \delta_{l_1,l_2}+ 
		\rho^{l_1+l_2}  m_{l_1+l_2}  \left(\frac{l_1+l_2}{2}\right)^{l_1+l_2} N^{(l_1+l_2)\left(\frac{1}{2}-\alpha\right) }.\\ 
		\end{array}
		\]
	\end{proof}
	A notable difference with the symmetric case already appears in the above lemma: the leading contribution is concentrated on the diagonal $l_1=l_2$. Consequently, the asymptotic moment matrix exhibits a much weaker coupling between different temporal scales. We can now precisely quantify the effect of increasing the reservoir dimension on the separation capacity.
	\begin{theorem}\label{Thm:NinftyGaussIID}
	Let $T\in \Nbb^*$, $\rho>0$ and $\alpha \in \Rbb$. For $N\in\Nbb^*$, let $W_N\in \Rbb^{N\times N}$ be a random matrix such that its entries are i.i.d.\ centred Gaussian random variables with standard deviation $\frac{\rho}{N^{\alpha}}$.
	\begin{itemize}
	\item If $\alpha=\frac12$, then 
	\[\begin{array}{ccl}
	\lambda_{\max}(B_{T,N}) 
	&\underset{N\to +\infty}{\simeq} &
	N\max(1,\rho^2,\ldots, \rho^{2T}),\\
	\lambda_{\min}(B_{T,N}) 
	&\underset{N\to +\infty}{\simeq} &
	N\min(1,\rho^2,\ldots, \rho^{2T}),\\
	\end{array}
	\]
	and
	\[
	\lim_{N\to \infty}\frac{\lambda_{\max}(B_{T,N})}{\sum\limits_{i=0}^T\lambda_{i} (B_{T,N})} =\frac{\max(1,\rho^2,\ldots, \rho^{2T})}{\sum\limits_{i=0}^T \rho^{2i}}.
	\]
	\item If $\alpha<\frac12$, then 
	\[
	\lambda_{\max}(B_{T,N}) 
	\underset{N\to +\infty}{\simeq} 
	\rho^{2T} N^{1+2T\left(\frac{1}{2}-\alpha\right)},
	\]
	and $\lambda_{\max}(B_{T,N})$ dominates the spectrum of $B_{T,N}$ as $N$ goes to infinity.
	\item If $\alpha>\frac12$, then 
	\[
	\lambda_{\max}(B_{T,N}) 
	\underset{N\to +\infty}{\simeq} 
	N,
	\]
	and $\lambda_{\max}(B_{T,N})$ dominates the spectrum of $B_{T,N}$ as $N$ goes to infinity.
	\end{itemize}
	\end{theorem}
	\begin{proof}
	For large $N$, Lemma \ref{lemma:OrderMagnitudeIID} implies 
	\[
	\begin{array}{ll}
		\sum\limits_{l=0}^T \rho^{2l} \frac{N(N-1)\cdots(N-l)}{N^{2 \alpha l}}  &\leq 
		\sum\limits_{l=0}^T B_{T,N}(l,l)\\
		& \leq 
		\sum\limits_{l=0}^T  \left( \rho^{2l}m_{2l} l^{2l} N^{2l\left(\frac{1}{2}-\alpha\right)} +
		  \rho^{2l} N^{1+2l\left(\frac{1}{2}-\alpha\right)} \right).\\
		\end{array}
	\]
	Hence
	\[
	\mathrm{tr}(B_{T,N}) 
	\underset{N\to +\infty}{\simeq}  
	\sum_{t=0}^T \rho^{2t} N^{1+2t\left(\frac{1}{2}-\alpha\right)}.
	\]
	Similarly
	\[
	\|B_{T,N}\|_{2,\infty}\underset{N\to +\infty}{\simeq}  
	\max_{0\leq t \leq T} \rho^{2t} N^{1+2t\left(\frac{1}{2}-\alpha\right)}.
	\]
	Like in the symmetric case, we have the following cases.
	
	\noindent\textbf{Case 1: }$\alpha<1/2$. Then (following Proposition \ref{prop:EVboundsNdim})
	\[
	\lambda_{\max}(B_{T,N}) 
	\underset{N\to +\infty}{\simeq} \rho^{2T} N^{1+2T\left(\frac{1}{2}-\alpha\right)},
	\]
	and $\lambda_{\max}(B_{T,N})$ dominates the spectrum of $B_{T,N}$ as $N$ goes to infinity.
	
	\noindent\textbf{Case 2: }$\alpha>1/2$. Then 
	\[
	\lambda_{\max}(B_{T,N}) 
	\underset{N\to +\infty}{\simeq} 
	N,
	\]
	and $\lambda_{\max}(B_{T,N})$ dominates the spectrum of $B_{T,N}$ as $N$ goes to infinity.
	
	\noindent\textbf{Case 3: }$\alpha=1/2$. By Lemma \ref{lemma:OrderMagnitudeIID}, one has
	\[
	\frac{B_{T,N}}{N} \underset{N\to +\infty}{\longrightarrow} 
	\mathrm{diag}(1,\rho^2,\ldots, \rho^{2T}),
	\]
	where $\mathrm{diag}(1,\rho^2,\ldots, \rho^{2T})$ is the $(T+1)\times(T+1)$ diagonal matrix whose diagonal is given by $(1,\rho^2,\ldots, \rho^{2T})$. This yields the claim.
	\end{proof}
	\begin{remark}\label{rem:RescaleUIID} As noted in the symmetric case, the extra factor $N$ appearing in Theorem \ref{Thm:NinftyGaussIID} is merely due to the norm of the chosen pre-processing vector $u$.
	\end{remark}
	Similarly to the symmetric case, Theorem \ref{Thm:NinftyGaussIID} shows an increase in the largest eigenvalue of $B_{T,N}$ that is at most polynomial in the dimension of the reservoir. We notice the same qualitative phase transition as in the symmetric setting: as $N\to\infty$, the critical scaling $\alpha=\frac12$ is the unique normalisation that prevents a deterioration in the quality of separation and a collapse of the spectrum onto a single dominant eigenvector as $N\to\infty$. We illustrate this in Figure \ref{Fig:GaussEVNNinftyIID} by plotting the evolution of dominance ratio $r_{T,N}$ as the reservoir dimension increases.
	
\begin{figure}[h!]
	\includegraphics[scale=0.4]{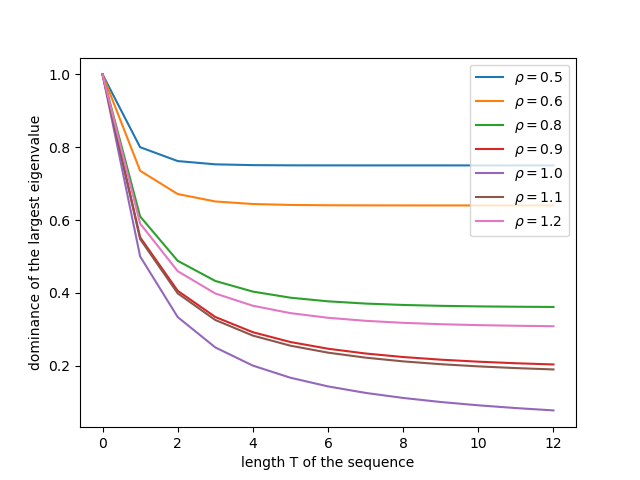}
	\centering
	\caption{The dominance of the largest eigenvalue, in function of the length $T$, over the entire spectrum of the limit in $N$ of the generalised matrix of moments $B_{T,N}$ associated to a random matrix with i.i.d.\ centred Gaussian entries with standard deviation $\rho/\sqrt{N}$.}
	\label{Fig:Lambda_dominance_dim_1_FCircle}
	\end{figure}

	Theorem \ref{Thm:NinftyGaussIID} also highlights that the limiting geometry is fundamentally different between the symmetric and the i.i.d.\ cases, even under the critical scaling $\alpha=\frac12$. While we obtain the rich moment structure of the semi-circle law in the limit of the symmetric case, the generalised matrix of moments in the i.i.d.\ case is asymptotically diagonal; resulting in a simple and explicit balance of separation across temporal directions. More particularly, separation is always most balanced and optimal for $\rho=1$, independently of the value of $T$, which sharply contrasts with the symmetric case. Figure~\ref{Fig:Lambda_dominance_dim_1_FCircle} shows the asymptotic dominance ratio predicted by Theorem~\ref{Thm:NinftyGaussIID}.
	
	We conclude this section by investigating the evolution of the quality of separation when the reservoir dimension remains fixed while the length of the input time series grows. We start with a simple comparative bound of the neighbouring entries of the matrix of moments $B_{T,N}$.
	\begin{lemma}\label{lemma:TinftyGaussIIDControl}
	Let $N\in \Nbb^*$ and $\sigma>0$. Let $W\in \Rbb^{N\times N}$ be a random matrix such that its entries are i.i.d.\ centred Gaussian random variables with standard deviation $\sigma$. For $T\in \Nbb^*$ and for all $l_1,l_2 \in \inti{0}{T-1}$, one has
	\begin{equation}\label{eq:SimpleBoundIID}
	B_{T,N}(l_1,l_2) \leq \frac{1}{\sigma^2N}B_{T,N}(l_1+1,l_2+1).
	\end{equation}
	Additionally, for all $l_1,l_2 \in \inti{0}{T-2}$
	\begin{equation}\label{eq:NotSimpleBoundIID}
	B_{T,N}(l_1,l_2) \leq \frac{1}{\sigma^4\left(\frac{l_1+l_2}{N^2}+1\right)}B_{T,N}(l_1+2,l_2+2).
	\end{equation}
	\end{lemma}
	\begin{proof} The statements are trivial when $l_1+l_2$ is odd. Assume then that $l_1+l_2$ is even. Let $\ibf=(i, i_1, \ldots, i_{l_1-1}, j) \in [N]^{l_1+1}$ and $\jbf=(i, j_1, \ldots, j_{l_2-1},j) \in [N]^{l_2+1}$ (note that $\ibf$ and $\jbf$ start and end with the same elements.) Then, for all $k\in [N]$, one has
	\[
	\sigma^2 \Ebb W_{\ibf}W_{\jbf} \leq  \Ebb W_{\ibf}W_{jk} W_{\jbf} W_{jk}.
	\]
	This implies 
	\[
	\sigma^2N \Ebb W_{\ibf}W_{\jbf} \leq \sum_{k=1}^N  \Ebb W_{\ibf}W_{jk} W_{\jbf} W_{jk},
	\]
	and therefore (summing over all pairs $(\ibf,\jbf)$ and using Identity \eqref{eq:WkWiWj})
	\[
	B_{T,N}(l_1,l_2) \leq \frac{1}{\sigma^2N}B_{T,N}(l_1+1,l_2+1).
	\]
	Consider now all the $l_1+l_2$ pairs in the following order
	\begin{equation}\label{eq:pairs}
	(i, i_1), (i_1, i_{2}), \ldots,
	(i_{l_1-1}, j), (i, j_1), (j_1, j_{2}), \ldots,
	(j_{l_2-1}, j).
	\end{equation}
	Let $k_{\ibf,\jbf}$ be the largest amount of times a pair appears in the above sequence \eqref{eq:pairs}, and let $(m,n)$ be the first pair appearing $k_{\ibf,\jbf}$ times in (\ref{eq:pairs}). For obvious reasons, let us only consider the cases where each pair in (\ref{eq:pairs}) appears an even number of times. Note that
	\[
	\frac{{l_1+l_2}}{N^2}
	\leq k_{\ibf,\jbf} \leq {l_1+l_2}.
	\] 
	We have then
	\[
	\sigma^4\left(\frac{l_1+l_2}{N^2}+1\right) \Ebb W_{\ibf}W_{\jbf}
	\leq 
	\sigma^4(k_{\ibf,\jbf}+1) \Ebb W_{\ibf}W_{\jbf}
	=\sigma^4 \frac{m_{k_{\ibf,\jbf}+2}}{m_{k_{\ibf,\jbf}}} \Ebb W_{\ibf}W_{\jbf} 
	\leq  \Ebb W_{\ibf}W_{jm}W_{mn} W_{\jbf} W_{jm}W_{mn}.
	\]
	Therefore 
	\[
	B_{T,N}(l_1,l_2) \leq \frac{1}{\sigma^4\left(\frac{l_1+l_2}{N^2}+1\right)}B_{T,N}(l_1+2,l_2+2).
	\]
	\end{proof}
	As a consequence of the above bounds, we have the following lower bounds on the dominance ratio $r_{T,N}$ in function of the length $T$ of the time series.
	\begin{theorem}\label{Thm:TinftyGaussIID}
	Let $N\in \Nbb^*$ and $\sigma>0$. Let $W\in \Rbb^{N\times N}$ be a random matrix such that its entries are i.i.d.\ centred Gaussian random variables with standard deviation $\sigma$. 
	Then, for all $T\in \Nbb$
	\[\frac{1}{P(T)}
	\leq 
	\frac{\lambda_{\max}(B_{T,N})}{\sum\limits_{i=0}^{T}\lambda_i(B_{T,N})}
	\leq 1,
	\] 
	where
	\[
	P(T)= \min\left(T+1,\sum_{l=0}^T(\sigma^2N)^{-l}\right).
	\]
	Moreover, there exists $T_{\sigma,N}>0$, such that, for all $T\geq T_{\sigma,N}$, one has
	\[
	\frac{1}{Q(T)}\leq 
	\frac{\lambda_{\max}(B_{T,N})}{\sum\limits_{i=0}^{T}\lambda_i(B_{T,N})},
	\]
	where
	\[
	Q(T):= 
	\min\left( 2,1+ \frac{1}{\sigma^2N} \right)
	\left( 1
	+ \frac{1}{2 \left(\frac{\sigma^2}{N}\right)^2(T-3)}
	+ \frac{T}{8 \left(\frac{\sigma^2}{N}\right)^4(T-5)^2}\right).
	\]
	\end{theorem}
	\begin{proof} A recursive argument using the bound \eqref{eq:SimpleBoundIID} in Lemma \ref{lemma:TinftyGaussIIDControl}, gives
	\[
	\mathrm{tr}(B_{T,N})\leq \sum_{l=0}^T \frac{1}{(\sigma^2N)^l} B_{T,N}(T,T).
	\]
	Thus
	\[
	\left( \sum_{l=0}^T \frac{1}{(\sigma^2N)^l}\right)^{-1}
	\leq 
	\frac{B_{T,N}(T,T)}{\mathrm{tr}(B_{T,N})}
	\leq 
	\frac{\lambda_{\max}(B_{T,N})}{\mathrm{tr}(B_{T,N})},
	\]
	which yields the first bound. Let $l\in \inti{0}{T}$. If $T-l$ is even and $T-l \geq 2$, the bound \eqref{eq:NotSimpleBoundIID} gives
	\[
	B_{T,N}(l,l) \leq \prod_{t=0}^{\frac{T-l}{2}-1}\frac{1}{\sigma^4\left(\frac{2(l+2t)}{N^2}+1\right)}B_{T,N}(T,T)
	=\prod_{k=1}^{\frac{T-l}{2}}\frac{1}{\sigma^4\left(\frac{2(T-2k)}{N^2}+1\right)}B_{T,N}(T,T).
	\]
	If $T-l$ is odd and $T-l\geq 3$, we similarly have
	\[
	B_{T,N}(l,l) \leq 
	\prod_{k=1}^{\frac{T-l-1}{2}}\frac{1}{\sigma^4\left(\frac{2(T-1-2k)}{N^2}+1\right)} B_{T,N}(T-1,T-1).
	\]
	We have then
	\[
	\sum_{T-l \text{ even}}B_{T,N}(l,l) \leq 
	\left(1+ \sum_{p=1}^{\lfloor T/2\rfloor} \prod_{k=1}^{p}\frac{1}{\sigma^4\left(\frac{2(T-2k)}{N^2}+1\right)}\right)B_{T,N}(T,T).
	\]
	Employing a similar technique than in the proof of Theorem \ref{Thm:TinftyGaussSym},  we can show that for $T$ larger than $T^{(1)}_{\sigma,N}$ (depending only on $\sigma$ and $N$), one has
	\[
	\sum_{T-l \text{ even}}B_{T,N}(l,l) \leq 
	\left(1
	+ \frac{1}{\sigma^4\left(\frac{2(T-2)}{N^2}+1\right)}
	+  \frac{1}{\sigma^8}\frac{\lfloor T/2\rfloor-1}{\left(\frac{2(T-2)}{N^2}+1\right)(\frac{2(T-4)}{N^2}+1)}\right)B_{T,N}(T,T).
	\]
	Similarly, for $T$ larger than $T^{(2)}_{\sigma,N}$ (depending only on $\sigma$ and $N$), one also has
	\[
	\sum_{T-l \text{ odd}}B_{T,N}(l,l) \leq 
	\left(1
	+ \frac{1}{\sigma^4\left(\frac{2(T-3)}{N^2}+1\right)}
	+  \frac{1}{\sigma^8}\frac{\lfloor \frac{T-1}{2} \rfloor-1}{\left(\frac{2(T-3)}{N^2}+1\right)(\frac{2(T-5)}{N^2}+1)}\right)B_{T,N}(T-1,T-1).
	\]
	Note that (using again the bound \eqref{eq:SimpleBoundIID})
	\[
	B_{T,N}(T,T) + B_{T,N}(T-1,T-1)
	\leq \min\left( 2,1+ \frac{1}{\sigma^2N} \right)
	\lambda_{\max}(B_{T,N}).
	\]
	Hence, defining 
	\[
	Q_0(T):= 1
	+ \frac{1}{2 \left(\frac{\sigma^2}{N}\right)^2(T-3)}
	+ \frac{T}{8 \left(\frac{\sigma^2}{N}\right)^4(T-5)^2},
	\]
	we get, for $T\geq T_{\sigma,N}:=\max\left(T^{(1)}_{\sigma,N}, T^{(2)}_{\sigma,N} \right)$
	\[
	\mathrm{tr}(B_{T,N}) \leq Q_0(T) (B_{T,N}(T,T) + B_{T,N}(T-1,T-1))
	\leq Q(T) \lambda_{\max}(B_{T,N}).
	\]
	This yields the result.
	\end{proof}
	
\begin{figure}[h!]
  \centering
  \begin{subfigure}{0.32\textwidth}
    \centering
    \adjustbox{valign=c}{\includegraphics[scale=0.33]{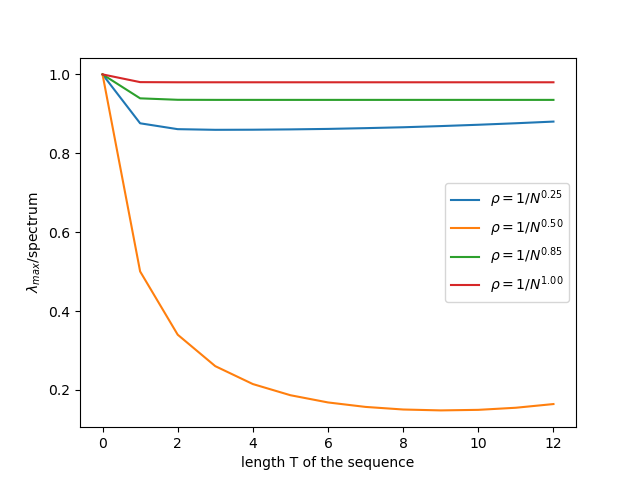}}
    \caption{$N=50$}
  \end{subfigure}
  \begin{subfigure}{0.32\textwidth}
    \centering
    \adjustbox{valign=c}{\includegraphics[scale=0.33]{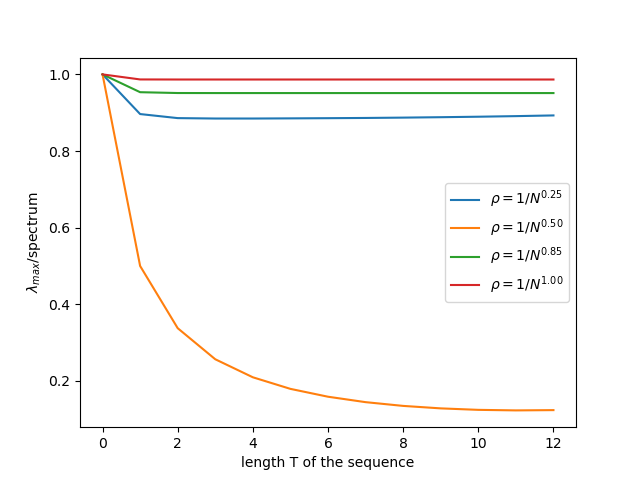}}
    \caption{$N=75$}
  \end{subfigure}
  \begin{subfigure}{0.32\textwidth}
    \centering
    \adjustbox{valign=c}{\includegraphics[scale=0.33]{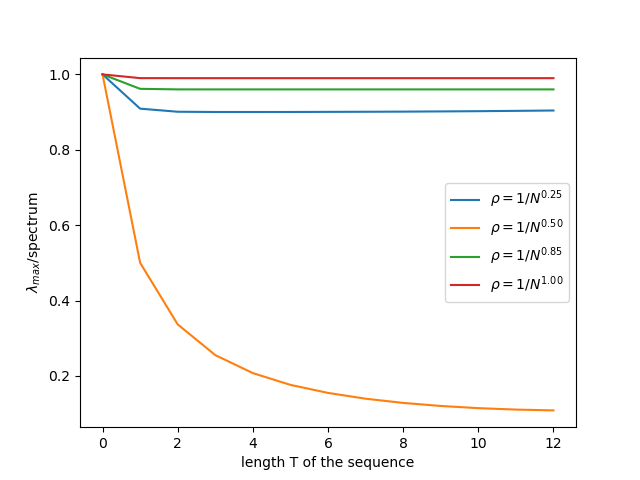}}
    \caption{$N=100$}
  \end{subfigure}
  \caption{The evolutions in function of the length of the time series $T$ of the dominance ratio $r_{T,N}$ (as defined in \eqref{eq:DomRatioN}) of the generalised matrix of moments associated to an $N\times N$ random connectivity matrix with i.i.d.\ centred Gaussian entries with standard deviation $\rho=\frac{1}{N^{\alpha}}$. The different panels correspond to different dimensions $N$ of the reservoir, while the different curves correspond to different values of the scaling exponent $\alpha$.}
  \label{Fig:GaussEVNTinftyIID}
  \end{figure}

	Unlike Theorem~\ref{Thm:TinftyGaussSym} in the symmetric setting, the result obtained in Theorem \ref{Thm:TinftyGaussIID} does not establish an unavoidable loss of separation quality for very long time series (understood as $(r_{T,N})_{T\geq 1}$ converging to $1$ as $T\to \infty$) in the i.i.d.\ Gaussian case. Whether such a collapse eventually occurs remains, at least from the perspective of these bounds, an open question. Nevertheless, the estimates clearly indicate that poor scalings (i.e.\ values of $\sigma$ far from the classical choice $N^{-\frac12}$) lead to a substantially larger spectral concentration and thus a deterioration in the quality of separation. We illustrate this behaviour in Figure \ref{Fig:GaussEVNTinftyIID} by plotting the dominance ratio $r_{T,N}$ as a function of $T$ corresponding to different values of the reservoir dimension $N$. As in the inequalities from Theorem \ref{Thm:TinftyGaussIID}, the plots point to a superiority of reservoir matrices having a standard deviation close to $N^{-\frac12}$, with the quality of separation improving with larger values of $N$.
	
	\begin{remark}\label{rem:TNbothinfty} More care is needed when taking both $T$ and $N$ going to infinity; in the sense that one needs to know at which rate both quantities grow. Indeed, considering the case of the scaling $\sigma=1/\sqrt{N}$, Theorem \ref{Thm:NinftyGaussIID} shows that, for all $T\geq 0$, one has
	\[
	\lim_{N\to +\infty} r_{T,N} =\frac{1}{T+1},
	\]
	Hence 
	\[
	\lim_{T\to +\infty}\left(\lim_{N\to +\infty} r_{T,N}\right) =0.
	\]
	However, the second bound from Theorem \ref{Thm:TinftyGaussIID} shows that, for every $N\in\Nbb^*$ and $\varepsilon>0$, for large values of $T$, one has $\frac{1}{2}-\varepsilon \leq r_{T,N}$.
	\end{remark}
\section{Illustrative numerical experiments}\label{sec:numerics}
	\subsection{General experimental setting}
	The objective of this section is to illustrate the practical relevance of the separation-based perspective developed in the previous sections, rather than establishing state-of-the-art performance on benchmark tasks. In particular, we investigate the relationship between the separation properties of random linear reservoirs and their empirical performance after training a downstream prediction model. To this end, we consider three representative learning tasks and evaluate families of reservoirs that differ in their dimension, normalisation, and connectivity structure. Although several design choices could considerably improve predictive performance (for example by introducing non-linear reservoir dynamics), our focus here is solely on understanding how the theoretical notions of separation introduced earlier manifest themselves in practice.

	\paragraph{Reservoir architecture.} More explicitly, and independently of the considered task, a scalar input sequence $\xbf=(x_t)_{0\leq t \leq T}$ is processed through a linear reservoir with pre-processing map $u$ and reservoir matrix $W$ following the recursion \eqref{eq:RCLinEq}. In particular, we take
	\[
	u=\frac1{\sqrt N}(1,\ldots,1)^\top.
	\]
	As discussed in Remarks \ref{rem:RescaleUSym} and \ref{rem:RescaleUIID}, this normalisation does not alter the qualitative separation properties studied in the theoretical sections. Its sole purpose is numerical, namely to prevent the reservoir states from growing excessively with the reservoir dimension. As for the reservoir matrix, we will consider both cases when $W$ is symmetric or has i.i.d.\ entries. The entries of $W$ will be centred Gaussian random variables with standard deviation $\frac{1}{N^{\alpha}}$, where $\alpha$ is a chosen parameter. We denote the sequence of reservoir states associated to the input time series $\xbf$ by $\zbf(\xbf)= (z_t(\xbf))_{0\leq t \leq T}$.

	\paragraph{Empirical measures of separation.} The theory developed in Section \ref{sec:ExpectDist} characterises separation through the spectrum of the matrix $B_{T,N}$. More precisely, the interval 
	\[[\lambda_{\min}(B_{T,N}), \lambda_{\max}(B_{T,N})]\] 
	and the concentration of the spectrum, quantified for instance by the dominance ratio $r_{T,N}$, provide task-independent and input-independent descriptions of the separation capacity of the reservoir. In practice, however, a given dataset only probes a restricted subset of the space of all possible time series. Consequently, the effective separation observed on a specific collection of inputs may differ from the worst-case or average behaviour predicted by the spectrum of $B_{T,N}$.

	To quantify this realised separation, we compute for each pair of distinct input sequences $\mathbf{x}$ and $\mathbf{y}$ the ratio
	\[
	r_{\xbf, \ybf}=\frac{\|\zbf(\xbf)-\zbf(\ybf)\|_2/\sqrt{N}}{\|\xbf-\ybf\|_2/\sqrt{T+1}},
	\]
	The use of normalised Euclidean norms is done for numerical stability and prevents the quantities involved from scaling artificially with either the reservoir dimension or the length of the time series. From these pairwise ratios, we derive four empirical proxies for the quality of separation
	\begin{equation}\label{eq:ProxySep}
	r_{\mathrm{m-m}}= \frac{\min_{\xbf\neq \ybf}r_{\xbf, \ybf}}{\max_{\xbf\neq \ybf} r_{\xbf, \ybf}},
	\;\; 
	r_{\mathrm{avg}}= \underset{\xbf\neq \ybf}{\mathrm{mean}} (r_{\xbf, \ybf}),
	\;\;
	r_{\mathrm{med}}= \underset{\xbf\neq \ybf}{\mathrm{median}} (r_{\xbf, \ybf})
	\;\text{and}\;
	r_{\mathrm{dom}}= \frac{\max_{\xbf\neq \ybf}r_{\xbf, \ybf}}{\sum_{\xbf\neq \ybf} r_{\xbf, \ybf}}.
	\end{equation}
	These quantities should be viewed as empirical counterparts of the spectral measures introduced in the theoretical part of the paper. More specifically
	\begin{itemize} 
	\item $r_{\mathrm{avg}}$ and $r_{\mathrm{med}}$ measure the typical level of separation achieved by the reservoir on the considered family of inputs.
	\item $r_{\mathrm{m-m}}$ quantifies the discrepancy in separation between the worst and best cases. 
	\item $r_{\mathrm{dom}}$ measures the extent to which separation is concentrated around a small number of particularly favourable directions.
	\item the discrepancy between $r_{\mathrm{avg}}$ and $r_{\mathrm{med}}$ provides additional information on the concentration properties of separation. 
	\end{itemize}
	A particularly desirable situation is the one highlighted by Theorem \ref{Thm:NinftyGaussIID} of equal separation in all directions, and which corresponds to the most popular practical setting (i.i.d.\ entries, $N$ large and $\alpha=\frac12$). From this perspective, poor separation quality is typically reflected by a very small value of $r_{\mathrm{m-m}}$, a large value of $r_{\mathrm{dom}}$ or a substantial difference between $r_{\mathrm{avg}}$ and $r_{\mathrm{med}}$. 
	
	It is important to emphasise that the quantities above remain entirely task-independent. They only depend on how the reservoir transforms the input signals and therefore provide information about its potential to support multiple downstream learning tasks. Task-specific notions of separation involving target outputs could also be considered, but this lies outside the scope of the present work.
	
	\paragraph{Readout architecture.} In order to accomplish the task in hand, a feedforward neural network $F$ with three hidden layers maps the reservoir states $(z_t(\xbf))_{0\leq t \leq T}$ to the desired outputs. The hidden layers of $F$ have $128$, $64$ and $16$ neurons respectively, each equipped with a ReLU activation function. The output layer and the loss function will depend on the task under consideration and will therefore be specified separately for each experiment.
	
	\paragraph{Experimental protocol.} Since the connectivity matrix is random, each experiment is repeated $p=50$ times using independently generated reservoirs. The separation measures defined in \eqref{eq:ProxySep}, as well as the corresponding training performance metrics (loss or accuracy), are then averaged across repetitions. Before discussing the numerical results, let us comment briefly on a few implementation choices.

The accompanying code allows the pre-processing vector $u$ to be trained jointly with the readout model. We deliberately disable this option in the experiments reported here. Indeed, as discussed in Remark~\ref{rem:GenUandD}, the separation properties of a reservoir depend on $u$, which would make comparisons between different reservoirs and simulations meaningless. Moreover, in several preliminary experiments, learning $u$ did not systematically improve the final performance.

The number of repetitions ($p=50$) may appear modest compared to the sample sizes often employed in Monte Carlo studies. Empirically, however, the averaged quantities stabilised after relatively few runs, and increasing $p$ further did not substantially modify our conclusions.

We also experimented with readout architectures whose width depended explicitly on the reservoir dimension $N$. Since this had only a limited impact on the results, we retained a fixed architecture across all experiments. This ensures that differences in performance can be more readily attributed to the reservoir itself rather than to changes in the expressive power of the readout network.
	\subsection{Classification: ECG5000}
	Our first case study concerns the classification of the ECG5000 dataset, a benchmark time-series classification problem originating from the UCR Time Series Archive. The dataset consists of electrocardiogram (ECG) recordings, each represented by a univariate time series of length $140$ and assigned to one of five heartbeat classes. The training dataset contains $n_{\mathrm{train}}=500$ time series, while the testing dataset contains $n_{\mathrm{test}}=4500$ times series.
	
	Before training, each signal is standardised using the mean and standard deviation computed from the training set. For a given input time series $\mathbf{x}$, we feed the final reservoir state $z_T(\mathbf{x})\in\Rbb^N$ to the readout network $F : \mathbb{R}^N \to \Rbb^5$, which maps this reservoir representation to the logits associated with the five heartbeat classes. The output layer is followed by a softmax transformation and the network is trained using the cross-entropy loss.
	
	For each experiment, we record both the separation measures introduced in \eqref{eq:ProxySep} and the classification accuracies on the training and test sets, denoted $\mathrm{acc}_{\mathrm{train}} $ and $\mathrm{acc}_{\mathrm{test}}$ respectively. All results are then averaged over the $p=50$ independent realisations of the random reservoir matrix.
	
	Table~\ref{tab:ECG_5000_alpha} reports the results obtained with a fixed reservoir dimension $N=100$ while varying the scaling exponent $\alpha$. We restricted ourselves to the contractive regime $\alpha\geq 0.6$ as, for smaller values of $\alpha$, both the reservoir states and the separation statistics become numerically unstable due to the rapid growth of matrix powers. We note first that the symmetric reservoir with $\alpha=0.6$ performs dramatically worse than all other architectures, which consistently achieve test accuracies around $90\%$. Secondly, this poor predictive performance is accompanied by a clear deterioration in the separation statistics. In particular, this reservoir exhibits the smallest value of $r_{\mathrm{m-m}}$ and the largest discrepancy between $r_{\mathrm{avg}}$ and $r_{\mathrm{med}}$. These observations are fully consistent with the point of view of separation discussed in this paper. Indeed, although the average amount of separation appears extremely large, separation is concentrated in only a few privileged directions while most directions remain poorly separated. Consequently, the resulting reservoir representation is highly unbalanced and appears ill-suited for classification in general. In contrast, all other reservoirs provide substantially more uniform separation, resulting in remarkably similar classification performance. This suggests that the overall quality of separation is more important than the absolute magnitude of separation itself.
	
\begin{table}[ht]
\centering
\begin{tabular}{c | c | c  c  c  c  c  c}
\hline
$\alpha$ & $W$ & $r_{\mathrm{m-m}}$ & $r_{\mathrm{avg}}$ & $r_{\mathrm{med}}$ & $r_{\mathrm{dom}}$ & $\mathrm{acc}_{\mathrm{train}}$ & $\mathrm{acc}_{\mathrm{test}}$ \\
\hline
 \multirow{2}{*}{$0.6$}  & sym & $2.13 \times 10^{-4}$ & $1.02 \times 10^{13}$ & $8.06 \times 10^{12}$ & $5.26 \times 10^{-5}$ & $47.03 \%$ & $46.52 \%$ \\
    & iid & $7.96 \times 10^{-3}$ & $3.09 \times 10^{-1}$ & $2.86 \times 10^{-1}$ & $2.69 \times 10^{-5}$ & $94.14 \%$ & $91.66 \%$ \\
\hline
   \multirow{2}{*}{$0.7$}  & sym & $6.66 \times 10^{-3}$ & $2.89 \times 10^{-1}$ & $2.62 \times 10^{-1}$ & $2.80 \times 10^{-5}$ & $93.35 \%$ & $91.70 \%$ \\
    & iid & $2.30 \times 10^{-3}$ & $2.38 \times 10^{-1}$ & $2.02 \times 10^{-1}$ & $3.42 \times 10^{-5}$ & $92.99 \%$ & $90.94 \%$ \\
\hline
 \multirow{2}{*}{$0.8$}  & sym & $1.28 \times 10^{-3}$ & $2.27 \times 10^{-1}$ & $1.90 \times 10^{-1}$ & $3.56 \times 10^{-5}$ & $92.59 \%$ & $90.55 \%$ \\
    & iid & $1.12 \times 10^{-3}$ & $2.15 \times 10^{-1}$ & $1.78 \times 10^{-1}$ & $3.77 \times 10^{-5}$ & $91.88 \%$ & $89.88 \% $ \\
\hline
\end{tabular}
\caption{ECG5000 classification results for reservoirs of dimension $N=100$. The table reports both classification performance and empirical separation statistics averaged over $50$ independent reservoir realisations.}
\label{tab:ECG_5000_alpha}
\end{table}	

	To further investigate this phenomenon, we fix $\alpha=0.6$ and vary the reservoir dimension. The results are reported in Table~\ref{tab:ECG_5000_N_var}. The most striking observation is that increasing the dimension of the symmetric reservoir fails to recover good classification performance. Despite dimensions as large as $N=300$, both training and test accuracies remain close to $50\%$. From the perspective of separation capacity, this behaviour is not at all surprising: the additional dimensions obtained from increasing $N$ do not generate a more balanced representation. Indeed, the values of $r_{\mathrm{m-m}}$ remain extremely small and the difference between $r_{\mathrm{avg}}$ and $r_{\mathrm{med}}$ remains substantial across all dimensions. The additional separation therefore concentrates along a small number of directions rather than being distributed throughout the representation space.
	
	The behaviour of reservoirs with i.i.d.\ connectivity matrices is markedly different: both the separation statistics and the classification accuracies remain remarkably stable across all tested dimensions. In particular, the quantity $r_{\mathrm{m-m}}$ remains approximately one order of magnitude larger than for the corresponding symmetric reservoirs, indicating a substantially more uniform distribution of separation across pairs of time series. Likewise, $r_{\mathrm{avg}}$ and $r_{\mathrm{med}}$ remain very close to one another and consistently lie on the same scale, suggesting that the observed separation is not dominated by a small number of atypical examples. At the same time, the test accuracy remains essentially constant, fluctuating only slightly around $91\%$ across all considered values of $N$.

Taken together, these results provide empirical motivation for one of the main themes of this paper: increasing the amount of separation alone is insufficient. What matters for (generic) downstream learning tasks is the ability of the reservoir to distribute separation across many directions of the input space rather than concentrating it into a few dominant modes.

\begin{table}[ht]
\centering
\begin{tabular}{c | c | c  c  c  c  c  c}
\hline
$N$ & $W$ & $r_{\mathrm{m-m}}$ & $r_{\mathrm{avg}}$ & $r_{\mathrm{med}}$ & $r_{\mathrm{dom}}$ & $\mathrm{acc}_{\mathrm{train}}$ & $\mathrm{acc}_{\mathrm{test}}$ \\
\hline
  \multirow{2}{*}{$50$} & sym & $1.35 \times 10^{-4}$ & $2.03 \times 10^{17}$ & $1.55 \times 10^{17}$ & $2.02 \times 10^{-4}$ & $48.70 \%$ & $48.01 \%$ \\
    & iid & $9.90 \times 10^{-3}$ & $4.54 \times 10^{-1}$ & $4.26 \times 10^{-1}$ & $2.64 \times 10^{-5}$ & $94.19 \%$ & $91.87 \%$ \\
\hline
  \multirow{2}{*}{$100$}  & sym & $2.13 \times 10^{-4}$ & $1.02 \times 10^{13}$ & $8.06 \times 10^{12}$ & $5.26 \times 10^{-5}$ & $47.03 \%$ & $46.52 \%$ \\
    & iid & $7.96 \times 10^{-3}$ & $3.09 \times 10^{-1}$ & $2.86 \times 10^{-1}$ & $2.69 \times 10^{-5}$ & $94.14 \%$ & $91.66 \%$ \\
\hline
  \multirow{2}{*}{$150$} & sym & $2.72 \times 10^{-4}$ & $9.94 \times 10^{9}$ & $8.71 \times 10^{9}$ & $4.81 \times 10^{-5}$ & $47.77 \%$ & $46.99 \%$ \\
    & iid & $6.46 \times 10^{-3}$ & $2.40 \times 10^{-1}$ & $2.19 \times 10^{-1}$ & $2.81 \times 10^{-5}$ & $94.05 \%$ & $91.58 \%$ \\
\hline
  \multirow{2}{*}{$200$} & sym & $2.96 \times 10^{-4}$ & $1.30 \times 10^{8}$ & $1.14 \times 10^{8}$ & $5.00 \times 10^{-5}$ & $49.49 \%$ & $48.57 \%$ \\
    & iid & $5.66 \times 10^{-3}$ & $2.01 \times 10^{-1}$ & $1.81 \times 10^{-1}$ & $2.90 \times 10^{-5}$ & $94.12 \%$ & $91.54 \%$ \\
\hline
  \multirow{2}{*}{$250$} & sym & $3.31 \times 10^{-4}$ & $1.94 \times 10^{7}$ & $1.75 \times 10^{7}$ & $3.93 \times 10^{-5}$ & $48.07 \%$ & $47.22 \%$ \\
    & iid & $5.24 \times 10^{-3}$ & $1.79 \times 10^{-1}$ & $1.61 \times 10^{-1}$ & $2.90 \times 10^{-5}$ & $94.11 \%$ & $91.57 \%$ \\
\hline
  \multirow{2}{*}{$300$}   & sym & $5.74 \times 10^{-4}$ & $4.11 \times 10^{5}$ & $3.48 \times 10^{5}$ & $4.57 \times 10^{-5}$ & $50.48 \%$ & $49.52 \%$ \\
    & iid & $4.90 \times 10^{-3}$ & $1.62 \times 10^{-1}$ & $1.45 \times 10^{-1}$ & $2.93 \times 10^{-5}$ & $94.06 \%$ & $91.51 \%$ \\
\hline
\end{tabular}
\caption{Influence of the reservoir dimension on the accuracy of the ECG5000 classification ($\alpha=0.6$).}
\label{tab:ECG_5000_N_var}
\end{table}
	\subsection{Memory: recalling the digits of $\pi$}
	Our second numerical experiment investigates a classical aspect of reservoir computing: memory. For this purpose, we consider a deterministic but highly irregular sequence and study the ability of the reservoir to retain information about its past inputs. More precisely, the input sequence $(x_t)_{0\leq t \leq T}$ is given by the first $T+1$ decimal digits of $\pi$. Given a fixed delay parameter $\tau=20$, the task consists of recovering the digit that entered the reservoir $\tau$ time steps earlier: for each time index $t\geq \tau$, the desired output is  $x_{t-\tau}$. 
	
	The reservoir state $z_t$ at time $t$ is processed by a feedforward neural network $F : \mathbb{R}^N \to \Delta^9$, where $\Delta^9$ denotes the probability simplex over the ten digit classes. The network outputs a probability vector $p_t=F(z_t)$ whose $i$-th component represents the predicted probability that the digit observed at time $t-\tau$ is equal to $i$. The target output is the one-hot vector $q_t=e_{x_{t-\tau}}\in\Rbb^{10}$, where $(e_0,\ldots,e_9)$ denotes the canonical basis of $\Rbb^{10}$.
	
	The available time indices $\mathcal{T} = \inti{\tau}{T}$ are split chronologically: the first $75\%$ of the indices are used for training and the remaining $25\%$ for testing. The readout network is trained using the cross-entropy loss and the Adam optimiser with learning rate $\eta= 10^{-3}$.
	
	Before discussing the numerical results, it is worth noting that this task admits an exact solution requiring neither learning nor random reservoirs. Indeed, by taking $u=(1,\ldots,1)^\top$, choosing the delay-line connectivity matrix
	\[
	W= \left( \begin{array}{cccccc}
	0&0&0&\cdots &0 &0\\
	1&0&0&\cdots &0 &0\\
	0&1&0&\cdots &0 &0\\
	\vdots & \vdots & \vdots & \ddots & \vdots & \vdots \\
	0&0&0&\cdots &1 &0\\
	\end{array}
	\right),
	\]
	and applying the following linear transformation on the reservoir states
	\[ 
	F\left(\begin{array}{c}
	z^{(0)}\\ z^{(1)} \\ \vdots \\ z^{({N-1})}\\
	\end{array}
	\right)=z^{({\tau})}-z^{({\tau-1})},
	\]
	then we can recover exactly the delayed digit. The purpose of the present experiment is therefore not to solve the memory problem optimally, but rather to compare random reservoirs of the types studied in this paper. In particular, we are interested in understanding whether the proposed separation measures correlate with the practical ability of a random reservoir to retain information about past inputs. More broadly, this experiment is related to the large body of work on memory capacity in reservoir computing (e.g.\ \cite{Jaeger, GGO3}). In particular, the general poor performance of random reservoirs in recalling past inputs (compared the delay-line and cyclic reservoirs) has been documented in \cite{VALT} through the numerical study of the rank of the controllability matrix $\Ccal$ (as defined in \eqref{eq:ControlMatrix}).
 
	Unlike the ECG5000 experiment, no genuine structure is expected to generalise beyond the training data. Indeed, the decimal expansion of $\pi$ is widely believed to behave similarly to a random sequence of digits. Consequently, the test accuracy should remain close to the baseline value of $10\%$ corresponding to random guessing among ten classes. From this perspective, the most informative quantity is the training accuracy, which measures the ability of the reservoir to preserve enough information about previous inputs for the readout network to memorise the delayed digits.

	The results obtained for a reservoir dimension $N=50$ are reported in Table~\ref{tab:pi_alpha}. For $\alpha=0.3$, we note that both symmetric and i.i.d.\ reservoirs perform almost identically and essentially fail to solve the task. The training accuracy remains close to $10\%$, indicating that the readout network is unable to extract meaningful information regarding the delayed digit. A striking difference appears at the critical scaling $\alpha=0.5$. While the symmetric reservoirs still perform close to random guessing, the i.i.d.\ reservoirs achieve a training accuracy exceeding $40\%$. This improvement is accompanied by a dramatic increase in $r_{\mathrm{m-m}}$ and a substantial decrease in $r_{\mathrm{dom}}$, indicating a much more balanced distribution of separation. When $\alpha=0.7$, both reservoir families exhibit non-trivial memory capacities coupled with good separation capacity profiles.

\begin{table}[ht]
\centering
\begin{tabular}{ c | c | c  c  c  c  c  c }
\hline
$\alpha$ & $W$ & $r_{\mathrm{m-m}}$ & $r_{\mathrm{avg}}$ & $r_{\mathrm{med}}$ & $r_{\mathrm{dom}}$ & $\mathrm{acc}_{\mathrm{train}}$ & $\mathrm{acc}_{\mathrm{test}}$ \\
\hline
  \multirow{2}{*}{$0.3$}  & sym & $3.67 \times 10^{-18}$ & $7.43 \times 10^{16}$ & $4.29 \times 10^{12}$ & $4.39 \times 10^{-2}$ & $9.52 \%$ & $5.63 \%$ \\
    & iid & $3.69 \times 10^{-18}$ & $5.84 \times 10^{16}$ & $4.22 \times 10^{12}$ & $1.62 \times 10^{-2}$ & $9.52 \%$ & $5.63 \%$ \\
\hline
  \multirow{2}{*}{$0.5$}  & sym & $1.98 \times 10^{-18}$ & $4.99 \times 10^{16}$ & $2.75 \times 10^{12}$ & $1.04 \times 10^{-2}$ & $9.52 \%$ & $5.63 \%$ \\
    & iid & $1.29 \times 10^{-2}$ & $4.26 \times 10^{15}$ & $3.11 \times 10^{11}$ & $3.35 \times 10^{-4}$ & $42.55 \%$ & $9.88 \%$ \\
\hline
  \multirow{2}{*}{$0.7$}  & sym & $2.78 \times 10^{-3}$ & $1.72 \times 10^{-1}$ & $1.48 \times 10^{-1}$ & $2.97 \times 10^{-4}$ & $37.25 \%$ & $14.25 \%$ \\
    & iid & $1.32 \times 10^{-3}$ & $1.59 \times 10^{-1}$ & $1.36 \times 10^{-1}$ & $3.13 \times 10^{-4}$ & $48.22 \%$ & $12.14\%$ \\
\hline
\end{tabular}
\caption{Performance statistics for random reservoirs on the delayed-digit recall task based on the decimal expansion of $\pi$ ($N=50$, $T=300$, $\tau=20$).}
\label{tab:pi_alpha}
\end{table}

	To investigate the asymptotic regime $N\to \infty$, we fix the critical scaling $\alpha=\frac12$ and vary the reservoir dimension. The results are reported in Table~\ref{tab:pi_N_var}. 

\begin{table}[ht]
\centering
\begin{tabular}{ c | c | c  c  c  c  c  c }
\hline
$N$ & $W$ & $r_{\mathrm{m-m}}$ & $r_{\mathrm{avg}}$ & $r_{\mathrm{med}}$ & $r_{\mathrm{dom}}$ & $\mathrm{acc}_{\mathrm{train}}$ & $\mathrm{acc}_{\mathrm{test}}$ \\
\hline
  \multirow{2}{*}{$25$}  & sym & $1.91 \times 10^{-18}$ & $4.00 \times 10^{16}$ & $2.23 \times 10^{12}$ & $9.41 \times 10^{-3}$ & $13.47 \%$ & $7.90 \%$ \\
    & iid & $3.23 \times 10^{-2}$ & $3.96 \times 10^{4}$ & $2.62 \times 10^{3}$ & $1.34 \times 10^{-3}$ & $76.27 \%$ & $8.57 \%$ \\
\hline
  \multirow{2}{*}{$50$}  & sym & $1.98 \times 10^{-18}$ & $2.89 \times 10^{16}$ & $1.59 \times 10^{12}$ & $1.04 \times 10^{-2}$ & $13.67 \%$ & $9.33 \%$ \\
    & iid & $5.04 \times 10^{-2}$ & $1.39 \times 10^{4}$ & $1.32 \times 10^{3}$ & $1.18 \times 10^{-3}$ & $79.27 \%$ & $8.85 \%$ \\
\hline
  \multirow{2}{*}{$100$}  & sym & $1.99 \times 10^{-18}$ & $2.01 \times 10^{16}$ & $9.98 \times 10^{11}$ & $1.11 \times 10^{-2}$ & $13.20 \%$ & $9.23 \%$ \\
    & iid & $6.89 \times 10^{-2}$ & $4.02 \times 10^{2}$ & $8.53 \times 10^{1}$ & $1.05 \times 10^{-3}$ & $86.10 \%$ & $8.95 \%$ \\
\hline
  \multirow{2}{*}{$250$}  & sym & $1.97 \times 10^{-18}$ & $1.34 \times 10^{16}$ & $6.51 \times 10^{11}$ & $1.13 \times 10^{-2}$ & $13.40 \%$ & $9.14 \%$ \\
    & iid & $1.60 \times 10^{-1}$ & $2.66 $ & $1.88 $ & $6.45 \times 10^{-4}$ & $98.43 \%$ & $9.14 \%$ \\
\hline
  \multirow{2}{*}{$500$} & sym & $2.00 \times 10^{-18}$ & $8.83 \times 10^{15}$ & $4.35 \times 10^{11}$ & $1.19 \times 10^{-2}$ & $13.27 \%$ & $8.66 \%$ \\
    & iid & $2.61 \times 10^{-1}$ & $5.72 \times 10^{-1}$ & $5.16 \times 10^{-1}$ & $4.56 \times 10^{-4}$ & $100 \%$ & $8.66 \%$ \\
\hline
\end{tabular}
\caption{
Influence of the reservoir dimension on the delayed-digit recall task for the critical scaling $\alpha=1/2$ ($T=100$).}
\label{tab:pi_N_var}
\end{table}
	The contrast between the two reservoir families is particularly striking. For symmetric reservoirs, the training accuracy remains close to random guessing for all tested values of $N$. Moreover, the empirical separation measures exhibit an extreme lack of uniformity: the quantity $r_{\mathrm{m-m}}$ remains essentially zero, while $r_{\mathrm{avg}}$ and $r_{\mathrm{med}}$ differ by several orders of magnitude. In contrast, the behaviour of the i.i.d.\ reservoirs improves steadily with increasing dimension. The quantity $r_{\mathrm{m-m}}$ grows consistently, the discrepancy between $r_{\mathrm{avg}}$ and $r_{\mathrm{med}}$ decreases, and the dominance score $r_{\mathrm{dom}}$ becomes progressively smaller. At the same time, the training accuracy increases from $76.27\%$ for $N=25$ to perfect memorisation at $N=500$. These observations are consistent with the theoretical results of Section~\ref{subsec:ExpIID}. The asymptotic regime characterised by Theorem~\ref{Thm:NinftyGaussIID} predicts increasingly balanced separation for properly scaled i.i.d.\ reservoirs as the dimension grows. The empirical results suggest that this balanced separation translates into a significantly enhanced ability to store and retrieve information about past inputs. 
 
 	While separation and memory are distinct notions, the present results suggest a strong practical connection between the two. In particular, for properly scaled random reservoirs, the ability to distribute separation across many directions appears closely related to the ability to store useful information over long temporal horizons.
	\subsection{Forecasting: the Lorenz dynamics}
	Our final experiment considers a forecasting task on one of the most classical examples of a chaotic dynamical system, namely the Lorenz system. The Lorenz system is defined by the autonomous differential equations
	\begin{equation}\label{eq:Lorenz}
  \dot{x}^{(1)} = \sigma(x^{(2)} - x^{(1)}),\qquad
  \dot{x}^{(2)} = x^{(1)}(\rho - x^{(3)}) - x^{(2)},\qquad
  \dot{x}^{(3)} = x^{(1)}x^{(2)} - \beta x^{(3)},
	\end{equation}
	with the classical parameter values $\sigma = 10$, $\rho = 28$, and $\beta = \tfrac{8}{3}$. A single Lorenz trajectory is generated by numerically integrating \eqref{eq:Lorenz} with a fourth-order Runge-Kutta scheme using the time step $\Delta t = 0.01$. The initial condition is drawn randomly once and then kept fixed throughout all experiments. This yields a three-dimensional time series $(x^{(1)}_t,x^{(2)}_t,x^{(3)}_t)_{0\leq t \leq T}$. Following a common setting in reservoir forecasting, only the first coordinate $(x^{(1)}_t)_{0 \leq t \leq T}$ is presented to the reservoir. For numerical stability, the inputs are standardised before being passed to the reservoir. 
	
	The forecasting task consists of predicting the entire next state of the Lorenz system. More precisely, at time $t\in\inti{0}{T-1}$, the reservoir receives $x^{(1)}_t$ (after standardisation) and is asked to predict the full next state $(x^{(1)}_{t+1}, x^{(2)}_{t+1}, x^{(3)}_{t+1})$. The readout network therefore implements a mapping $F:\Rbb^N\to\Rbb^3$. Its output layer consists of a linear transformation with no activation function.
	
	As in the previous experiment, the time indices are split chronologically: the first $75\%$ of the available observations are used for training and the remaining $25\%$ for testing. Training is performed using the mean squared error (MSE) loss. The trainable parameters are optimised using Adam with learning rate $\eta=10^{-3}$ over $300$ epochs.
	
	We again restrict ourselves to the contractive regime $\alpha\geq 0.6$ since smaller values result in severe numerical instabilities caused by the rapid growth of matrix powers. The results obtained for reservoirs of dimension $N=150$ and time horizon $T=150$ are reported in Table~\ref{tab:lorenz_alpha}. The most striking observation is the behaviour of the symmetric reservoir when $\alpha=0.6$. Both the training and test losses are extremely large, indicating that the reservoir dynamics are close to a numerical blow-up regime. This phenomenon is reflected very clearly in the separation statistics: the average separation score reaches values of order $10^{9}$, while $r_{\mathrm{m-m}}$ becomes essentially zero and the discrepancy between $r_{\mathrm{avg}}$ and $r_{\mathrm{med}}$ spans several orders of magnitude. From the perspective developed throughout this paper, the explanation is natural. The reservoir does not merely separate inputs strongly; it separates a very small subset of input pairs at several orders of magnitudes more than all others. As a consequence, the resulting representation is highly ill-conditioned and difficult for the readout network to exploit. Once $\alpha$ is increased to $0.7$ or $0.8$, the dynamics become stable and the two reservoir families exhibit much more comparable behaviour. In this regime, the separation statistics are of similar magnitude and the train and test losses become comparable as well. This suggests that the elimination of extreme spectral concentration leads to representations that are substantially easier to learn from.

\begin{table}[ht]
\centering
\begin{tabular}{ c | c | c  c  c  c  c  c }
\hline
$\alpha$ & $W$ & $r_{\mathrm{m-m}}$ & $r_{\mathrm{avg}}$ & $r_{\mathrm{med}}$ & $r_{\mathrm{dom}}$ & $\mathcal{L}_{\mathrm{train}}$ & $\mathcal{L}_{\mathrm{test}}$ \\
\hline
  \multirow{2}{*}{$0.6$}  & sym & $3.90 \times 10^{-10}$ & $7.80 \times 10^{9}$ & $1.49 \times 10^{7}$ & $6.36 \times 10^{-3}$ & $3.27 \times 10^{5}$ & $1.10 \times 10^{15}$ \\
    & iid & $7.48 \times 10^{-4}$ & $1.16 \times 10^{-1}$ & $9.65 \times 10^{-2}$ & $4.69 \times 10^{-4}$ & $29.35$ & $11.42$ \\
\hline
  \multirow{2}{*}{$0.7$}  & sym & $1.21 \times 10^{-3}$ & $1.19 \times 10^{-1}$ & $9.98 \times 10^{-2}$ & $2.85 \times 10^{-4}$ & $33.19$ & $11.52$ \\
    & iid & $5.78 \times 10^{-4}$ & $9.86 \times 10^{-2}$ & $8.13 \times 10^{-2}$ & $3.48 \times 10^{-4}$ & $36.65$ & $10.42$ \\
\hline
  \multirow{2}{*}{$0.8$}  & sym & $7.00 \times 10^{-4}$ & $9.90 \times 10^{-2}$ & $8.17 \times 10^{-2}$ & $2.50 \times 10^{-4}$ & $38.76$ & $10.78$ \\
    & iid & $5.90 \times 10^{-4}$ & $9.39 \times 10^{-2}$ & $7.73 \times 10^{-2}$ & $2.57 \times 10^{-4}$ & $38.38 $ & $10.29$ \\
\hline
\end{tabular}
\caption{Performance metrics of random reservoirs (of dimension $N=150$) for the task of one-step-ahead forecasting of the Lorenz dynamics based on the first current Lorenz coordinate (with time horizon $T=150$.)}
\label{tab:lorenz_alpha}
\end{table}

	To further investigate this instability, we now fix $\alpha=0.6$ and vary the reservoir dimension. The results are shown in Table~\ref{tab:lorenz_N_var}.  The behaviour of the i.i.d.\ reservoirs is remarkably stable. Across all tested dimensions, the separation statistics evolve only moderately and both the training and test losses remain within the same order of magnitude. In particular, the quantities $r_{\mathrm{m-m}}$ and $r_{\mathrm{dom}}$ remain almost constant while the difference between $r_{\mathrm{avg}}$ and $r_{\mathrm{med}}$ remains limited. The situation is very different for symmetric reservoirs. For small and moderate dimensions, the reservoir operates in a regime of extreme separation imbalance. Training and test losses are sometimes so large that meaningful forecasting becomes impossible. As the dimension increases, however, the separation profile gradually improves: $r_{\mathrm{m-m}}$ increases by several orders of magnitude, $r_{\mathrm{dom}}$ decreases, and the discrepancy between $r_{\mathrm{avg}}$ and $r_{\mathrm{med}}$ becomes less pronounced. This improvement is accompanied by a substantial reduction in training loss. For example, the mean training loss drops from numerical values of order $10^{32}$ for $N=25$ to values of order unity when $N=500$. Although the corresponding test losses remain significantly larger than those of the i.i.d.\ reservoirs, the overall trend strongly suggests that more balanced separation is associated with more effective learning of the underlying dynamics.
\begin{table}[ht]
\centering
\begin{tabular}{c | c | c  c  c  c  c  c }
\hline
$N$ & $W$ & $r_{\mathrm{m-m}}$ & $r_{\mathrm{avg}}$ & $r_{\mathrm{med}}$ & $r_{\mathrm{dom}}$ & $\mathcal{L}_{\mathrm{train}}$ & $\mathcal{L}_{\mathrm{test}}$ \\
\hline
  \multirow{2}{*}{$25$}   & sym & $5.31 \times 10^{-12}$ & $6.24 \times 10^{16}$ & $2.44 \times 10^{12}$ & $1.23 \times 10^{-2}$ & $3.57 \times 10^{32}$ & $\infty$ \\
    & iid & $8.56 \times 10^{-4}$ & $3.37 \times 10^{-1}$ & $2.85 \times 10^{-1}$ & $4.96 \times 10^{-4}$ & $26.58$ & $9.63$ \\
\hline
  \multirow{2}{*}{$50$}   & sym & $1.12 \times 10^{-12}$ & $2.02 \times 10^{16}$ & $8.92 \times 10^{11}$ & $9.92 \times 10^{-3}$ & $4.31 \times 10^{23}$ & $1.29 \times 10^{35}$ \\
    & iid & $8.25 \times 10^{-4}$ & $2.16 \times 10^{-1}$ & $1.82 \times 10^{-1}$ & $4.92 \times 10^{-4}$ & $27.15 $ & $10.03 $ \\
\hline
  \multirow{2}{*}{$100$}   & sym & $4.99 \times 10^{-11}$ & $1.41 \times 10^{13}$ & $1.10 \times 10^{9}$ & $8.67 \times 10^{-3}$ & $9.43 \times 10^{11}$ & $1.06 \times 10^{22}$ \\
    & iid & $7.99 \times 10^{-4}$ & $1.49 \times 10^{-1}$ & $1.24 \times 10^{-1}$ & $4.67 \times 10^{-4}$ & $27.96$ & $10.73$ \\
\hline
  \multirow{2}{*}{$150$}   & sym & $3.90 \times 10^{-10}$ & $7.80 \times 10^{9}$ & $1.49 \times 10^{7}$ & $6.36 \times 10^{-3}$ & $3.27 \times 10^{5}$ & $1.10 \times 10^{15}$ \\
    & iid & $7.48 \times 10^{-4}$ & $1.16 \times 10^{-1}$ & $9.65 \times 10^{-2}$ & $4.69 \times 10^{-4}$ & $29.35$ & $11.42$ \\
\hline
  \multirow{2}{*}{$200$}   & sym & $6.62 \times 10^{-9}$ & $8.85 \times 10^{7}$ & $6.07 \times 10^{5}$ & $5.76 \times 10^{-3}$ & $7.04 \times 10^{4}$ & $1.28 \times 10^{13}$ \\
    & iid & $7.13 \times 10^{-4}$ & $9.81 \times 10^{-2}$ & $8.13 \times 10^{-2}$ & $4.64 \times 10^{-4}$ & $29.87$ & $11.77$ \\
\hline
  \multirow{2}{*}{$250$} & sym & $9.28 \times 10^{-8}$ & $1.26 \times 10^{7}$ & $1.30 \times 10^{5}$ & $3.05 \times 10^{-3}$ & $1.41 \times 10^{2}$ & $1.44 \times 10^{8}$ \\
    & iid & $7.14 \times 10^{-4}$ & $8.74 \times 10^{-2}$ & $7.25 \times 10^{-2}$ & $4.58 \times 10^{-4}$ & $31.50$ & $12.61$ \\
\hline
  \multirow{2}{*}{$500$}  & sym & $3.19 \times 10^{-4}$ & $1.50 \times 10^{2}$ & $3.23 \times 10^{1}$ & $1.38 \times 10^{-3}$ & $6.29 $ & $1.197 \times 10^{4}$ \\
    & iid & $6.53 \times 10^{-4}$ & $5.95 \times 10^{-2}$ & $4.92 \times 10^{-2}$ & $4.48 \times 10^{-4}$ & $32.74$ & $13.83$ \\
\hline
\end{tabular}
\caption{Influence of the reservoir dimension on Lorenz forecasting for the scaling $\alpha=0.6$ (with time horizon $T=150$).}
\label{tab:lorenz_N_var}
\end{table}

	In summary, and while the observations following the three experimental studies presented in this section do not necessarily establish a causal relationship, they provide empirical evidence that the geometric quantities introduced in this paper capture meaningful aspects of reservoir quality. In particular, the results consistently suggest that the most useful reservoirs are not those that maximise separation in a few privileged directions, but rather those that distribute separation more uniformly across the space of inputs.
\section{Concluding remarks and open problems}\label{sec:OProblems}
The objective of this work was to develop a mathematical framework for understanding the separation capacity of random linear reservoirs. Our main observation is that, in expectation, separation is completely encoded by the spectrum of a positive semi-definite matrix naturally associated with the connectivity matrix of the reservoir, namely the generalised matrix of moments $B_{T,N}$. In particular, we showed that the expected distortion of distances generated by a linear reservoir can be characterised through the eigenvalues of $B_{T,N}$ and argued that the concentration of its spectrum plays a decisive role. For both symmetric and i.i.d.\ Gaussian connectivity matrices, we obtained explicit asymptotic descriptions of the dominant eigenvalues and identified regimes in which separation is either distributed across many directions or concentrated along the dominant eigenvector.

	The numerical experiments further suggest that these spectral quantities capture practically relevant aspects of reservoir quality. Across classification, memory, and forecasting tasks, reservoirs exhibiting a more balanced separation profile consistently produced more useful representations than reservoirs whose separation was concentrated in a small number of directions. While these observations remain empirical, they provide encouraging evidence that separation capacity may serve as a meaningful complement to more classical notions such as memory capacity.

	The present work naturally raises a number of theoretical and practical questions. For instance, throughout the paper, our analysis focused primarily on the eigenvalues of the matrix $B_{T,N}$. A natural continuation is to investigate its eigenvectors and their evolution with the time horizon $T$. Such a study would reveal which temporal directions are preferentially separated by a given reservoir and would provide a finer description of the geometry of the induced representation. More specifically, one may analyse the localisation of the eigenvectors of $B_{T,N}$ in the canonical temporal basis $((\delta_{i,j})_{0\leq j\leq T})_{0\leq i\leq T}$, thereby identifying whether separation is driven primarily by recent inputs, distant inputs, or a combination of both. For deterministic cyclic reservoirs, related questions were investigated in \cite{FLT}, where an intriguing connection with Fourier decompositions of the input signal was uncovered. More generally, understanding the geometry of these eigenvectors could shed light on some of the limitations of classical reservoir computing when compared with modern architectures based on attention mechanisms, where relevant information from arbitrary locations in a sequence can be accessed directly.
	
	Another question concerns the degradation of separation for long time series in the i.i.d.\ case. While Theorem \ref{Thm:TinftyGaussIID} partially addresses this issue for poorly scaled reservoirs, one is tempted to conjecture that this separation capacity should still deteriorate, perhaps at a different rate, as $T\to \infty$, since the collection of reservoir states is ultimately constrained to live in the finite-dimensional space $\Rbb^N$. Determining whether such a collapse indeed occurs, and obtaining sharp asymptotic rates, remains an open problem. Moreover, the simultaneous limit $N,T\to\infty$ appears considerably more delicate than either limit taken separately.
	
	In light of the results of Section \ref{sec:ExpectDist}, the natural next step is to understand to what extent expected separation translates into separation with high probability. More precisely, one would like to obtain inequalities of the form
		\[
	\Pbb\left( \|f(\xbf,W)-f(\ybf,W)\|^2_2
	\geq M \|\xbf-\ybf\|^2_2
	\right)\geq 1- \delta,
	\]
	for suitably large $M>0$ and small $\delta>0$, and ideally uniformly over all admissible input time series. In the linear setting considered here, $\|f(\xbf,W)-f(\ybf,W)\|^2_2$ is a polynomial in independent random variables. Consequently, one may attempt to use concentration inequalities for polynomial chaoses, such as those developed in \cite{AW}. However, several difficulties arise. First, polynomial functions typically exhibit heavier tails than linear statistics. Second, these theoretical tails depend on a large collection of data-dependent tensor norms whose practical evaluation can be prohibitively difficult. Finally, the resulting estimates are generally quite conservative and may fail to capture the true behaviour observed in practice. Nevertheless, obtaining even qualitative high-probability guarantees would significantly strengthen the theory developed in this paper.
	
	From a theoretical standpoint, perhaps the most important extension is the incorporation of non-linear reservoir dynamics. We expect that suitable non-linearities, such as those encountered in practice, introduce a more consistent separation behaviour (i.e.\ less discrepancy between expected and high-probability separation.) A key challenge here, for high-probability bounds, is the lack of two-sided probability concentration inequalities as those readily available for polynomial functions in Gaussian random variables.
	
	The present work also suggests new approaches for reservoir design. For example, rather than selecting hyperparameters solely through downstream training performance, one may attempt to optimise the separation capacity of a reservoir directly, either in a data- or task-dependent manner. Such an approach would partially decouple hyperparameter tuning from the subsequent task-dependent training of the readout layer and could potentially lead to more robust and efficient RC systems. This perspective is closely related to previous work on memory-capacity optimisation (\cite{OPG}, although \cite{Marzen} shows that optimising the memory capacity of linear networks does not necessarily translate into a good forecasting capacity.)
	
	Throughout this article, we studied separation uniformly over arbitrary input time series. In many practical situations, however, only a specific class of inputs is relevant. This suggests considering separation capacity relative to a prescribed input distribution, much in the same spirit as memory-capacity analyses for i.i.d.\ or stationary inputs (\cite{Jaeger,GGO3}). Maximising separation on a target family of signals may be considerably more effective than maximising separation uniformly over the entire space of time series. Additionally, it would be interesting on the one hand to directly compare the notions of separation and memory capacity (beyond the comments made in Remark \ref{rem:LinkMemoryControl}), and, on the other hand, to link separation capacity to generalisation error bounds.
	\bibliographystyle{alpha}
	\bibliography{bibRRC}
\end{document}